\documentclass[letterpaper]{article} 
\usepackage{aaai2026}  
\nocopyright
\usepackage{times}  
\usepackage{helvet}  
\usepackage{courier}  
\usepackage[hyphens]{url}  
\usepackage{graphicx} 
\usepackage{natbib}  
\usepackage{caption} 
\usepackage{algorithm}
\usepackage{algorithmic}

\usepackage{newfloat}
\usepackage{listings}
\DeclareCaptionStyle{ruled}{labelfont=normalfont,labelsep=colon,strut=off} 
\floatstyle{ruled}
\newfloat{listing}{tb}{lst}{}
\floatname{listing}{Listing}
\usepackage{amsmath}	
\usepackage{amssymb}
\usepackage{cleveref}
\usepackage{multirow}
\usepackage{tabularx}
\usepackage{subcaption}

\title{MUJICA: Reforming SISR Models for PBR Material Super-Resolution via Cross-Map Attention}
\author {
    Xin Du\textsuperscript{\rm 1},
    Maoyuan Xu\textsuperscript{\rm 1}\thanks{Corresponding author.},
    Zhi Ying\textsuperscript{\rm 1}
}
\affiliations {
    \textsuperscript{\rm 1}Ubisoft La Forge\\
    goldenbiglord@gmail.com, mao-yuan.xu@ubisoft.com, zhi.ying@ubisoft.com
}

\usepackage{bibentry}

\begin{document}


\maketitle

\begin{abstract}
Physically Based Rendering (PBR) materials are typically characterized by multiple 2D texture maps such as basecolor, normal, metallic, and roughness which encode spatially-varying bi-directional reflectance distribution function (SVBRDF) parameters to model surface reflectance properties and microfacet interactions. Upscaling SVBRDF material is valuable for modern 3D graphics applications. However, existing Single Image Super-Resolution (SISR) methods struggle with cross-map inconsistency, inadequate modeling of modality-specific features, and limited generalization due to data distribution shifts. In this work, we propose \textbf{M}ulti-modal \textbf{U}pscaling \textbf{J}oint \textbf{I}nference via \textbf{C}ross-map \textbf{A}ttention (\textbf{MUJICA}), a flexible adapter that reforms pre-trained Swin-transformer-based SISR models for PBR material super-resolution. MUJICA is seamlessly attached after the pre-trained and frozen SISR backbone. It leverages cross-map attention to fuse features while preserving remarkable reconstruction ability of the pre-trained SISR model. Applied to SISR models such as SwinIR, DRCT and HMANet, MUJICA improves PSNR, SSIM, and LPIPS scores while preserving cross-map consistency. Experiments demonstrate that MUJICA enables efficient training even with limited resources and delivers state-of-the-art performance on PBR material datasets.
\end{abstract}


\section{Introduction}

\label{sec:intro}
Physically Based Rendering (PBR) materials consist of different 2D texture maps such as basecolor, normal, roughness and metallic, that describe the appearance of virtual 3D shapes under arbitrary lighting in the form of spatially-varying bi-directional reflectance distribution function (SVBRDF). The industry continues to explore ways to integrate deep learning methods into modern PBR material production pipeline for various reasons. One major motivation is the requirement to upscale low-resolution legacy PBR materials to a higher resolution when developing remastered versions of games. Since legacy assets were created without using tools like Adobe Substance 3D Sampler, it is usually not possible to regenerate at arbitrary resolution. 
Another key factor is the growing adoption of diffusion models~\cite{podell2023sdxl,ho2020denoising,sohl2015deep,song2019generative,song2020score} to generate high-quality PBR materials~\cite{vecchio2024controlmat,ye2024stablenormal,saravanan2025generative,vecchio2024matfuse,esser2024scaling}. However, the native output resolution of these models is typically limited to $512 \times 512$ or $1024 \times 1024$ in current industrial practice due to limited computation resources, failing to reach industrial requirements such as $4096 \times 4096$. Therefore, reconstructing high-resolution PBR materials from low-resolution inputs has become an important component of modern industry pipelines.

\begin{figure}[htb]
\centering
    \includegraphics[width=1\linewidth]{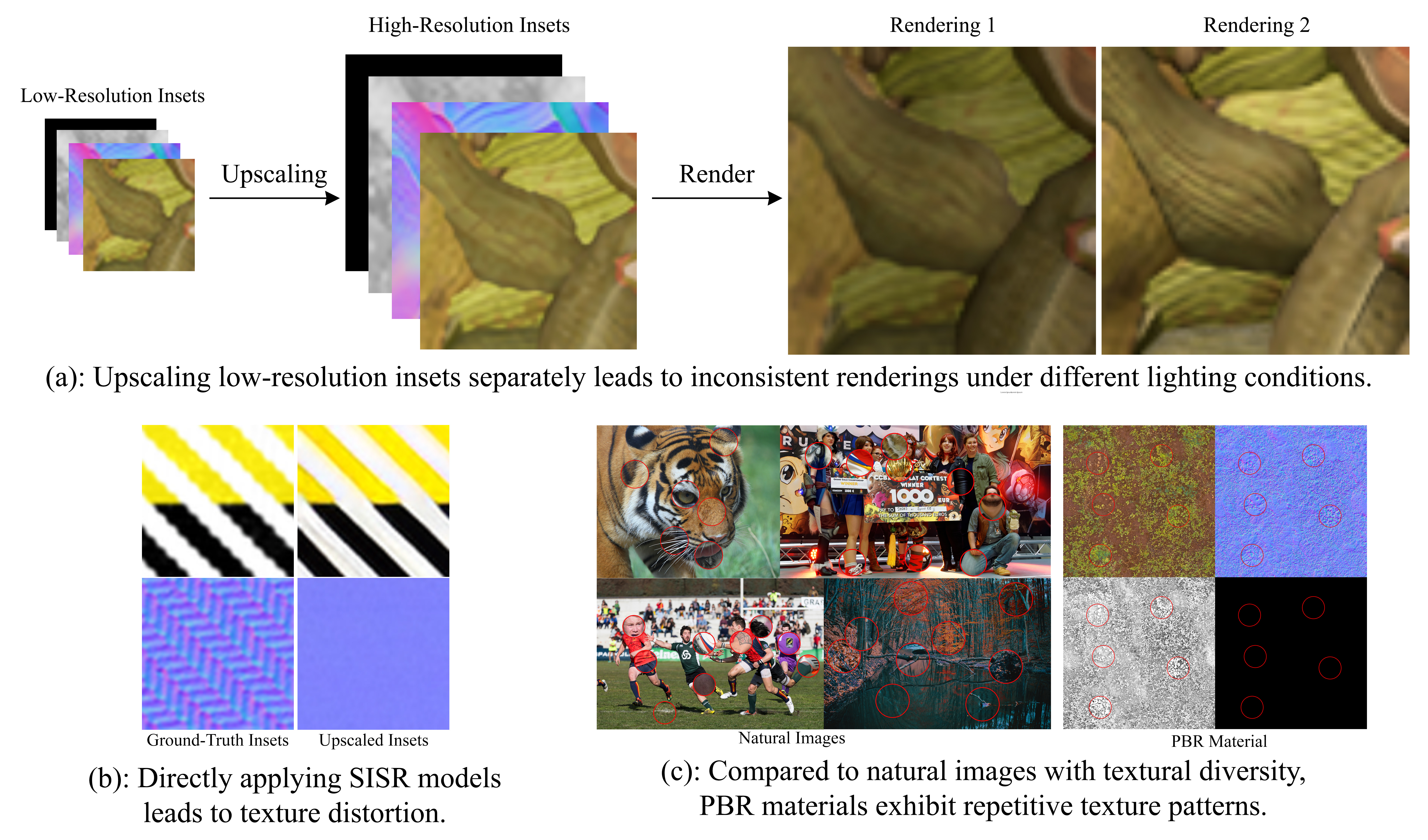}
\caption{Limitations of directly applying SISR models on PBR materials.}
\label{fig:pbr_challenges}
\end{figure}

In the task of Single Image Super-Resolution (SISR), methods based on Swin Transformer~\cite{liu2021swin} have achieved remarkable success over CNN-based methods~\cite{8099502,Lim_2017_CVPR_Workshops,NEURIPS2020_eaae339c,wang2018esrgan,zhang2018image}, but their direct application to PBR materials faces three main limitations: 
\begin{itemize}
    \item \textbf{Cross-Map Inconsistency.} As shown in \cref{fig:pbr_challenges} (a), SISR models process each PBR material map such as basecolor and normal separately, breaking consistency between maps. Such map inconsistency results in inconsistent renderings under varying lighting conditions.
    \item \textbf{Texture Distortion.} \Cref{fig:pbr_challenges} (b) demonstrates that directly applying pre-trained state-of-the-art (SOTA) SISR models to PBR materials produces incorrect results like texture distortion. This is mostly due to the different data distribution between the natural images that SISR models are trained on and the PBR materials.
    \item \textbf{Limited Datasets.} Available datasets for training PBR material super-resolution (SR) models are much more limited compared with those for SISR. To train our model, we employ MatSynth~\cite{vecchio2024matsynth}, the only publicly available SVBRDF dataset for our task, along with our in-house dataset. Remarkably, the textural and compositional complexity in single natural image often significantly surpasses single PBR material as shown in \cref{fig:pbr_challenges} (c). Hence, training a PBR material SR model with limited data resources presents a significant challenge.
\end{itemize}

To address the challenges above, we propose \textbf{M}ulti-modal \textbf{U}pscaling \textbf{J}oint \textbf{I}nference via \textbf{C}ross-map \textbf{A}ttention (\textbf{MUJICA}) in this paper to adapt SOTA SISR models like SwinIR~\cite{9607618}, DRCT~\cite{Hsu_2024_CVPR} and HMANet~\cite{Chu_2024_CVPR} to PBR material SR task by cross-map feature fusion. To better capture the intricate interdependence among different PBR material maps, MUJICA adopts an attention-based fusion strategy. This design allows for both shared representation learning and flexible map specific refinement, facilitating robust multi-modal modeling under the complex structural priors inherent in PBR materials. Our contributions are summarized as below
\begin{itemize}
    \item \textbf{Multi-modal Upscaling.} MUJICA reformulates PBR material SR as a multi‑modal fusion problem, achieving SOTA performance on existing PBR material datasets.
    \item \textbf{Adapter for SISR models.} MUJICA as an adapter to reform existing Swin‑transformer‑based SISR models to multi-modal SR models. Equipped with MUJICA, existing SISR models have the ability to deal with PBR materials while keeping their remarkable performance.
    \item \textbf{Efficient Training.} By integrating MUJICA with any frozen, pre‑trained Swin‑transformer‑based SISR model as the backbone, only a small number of parameters need to be trained. Therefore, our proposed method can keep training requirements and hardware demands at a relatively low level while achieving SOTA performance even with limited training data.
\end{itemize}

Experiments demonstrate that MUJICA achieves SOTA performance on existing PBR material datasets. For $\times 2$ SR task, MUJICAs outperform their SISR backbones with gains of up to \textbf{1.15dB} in PSNR, \textbf{0.0069} in SSIM, and a reduction of \textbf{0.036} in LPIPS on renderings across datasets. For $\times 4$ SR task, MUJICAs improve metrics up to \textbf{0.76dB} in PSNR, \textbf{0.0070} in SSIM, and \textbf{0.0695} in LPIPS.

\section{Related Work}

\label{sec:related}
\subsection{Single Image Super-Resolution}
\begin{figure}[htb]
\centering
    \includegraphics[width=1\linewidth]{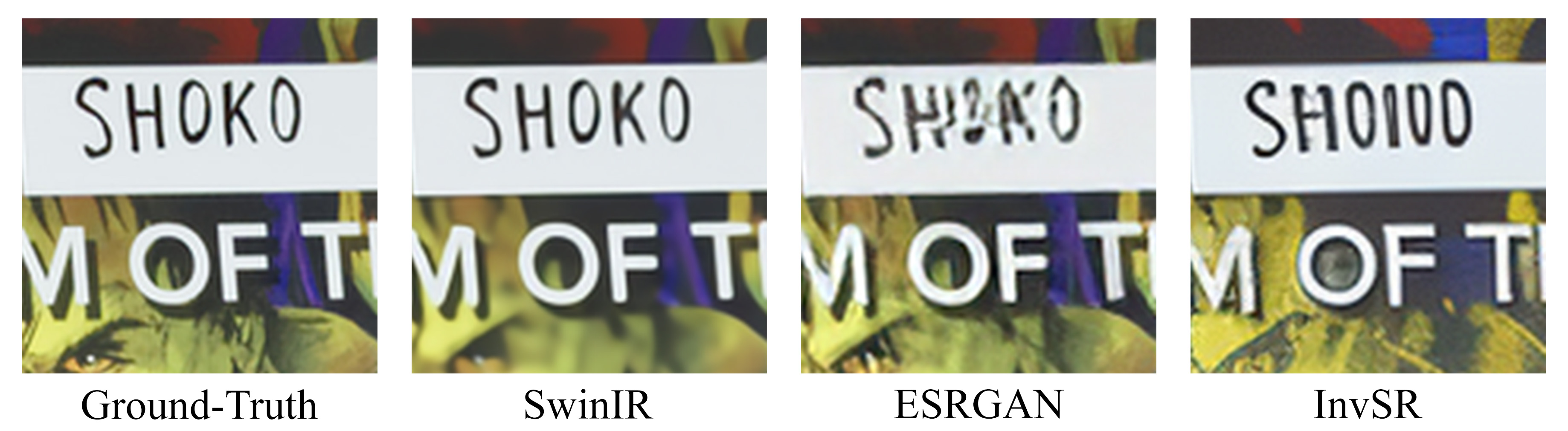}
\caption{Visual comparison on a natural image inset. ESRGAN and InvSR produce perceptually sharper textures but introduce structural distortions, while Swin-transformer-based method demonstrates superior capability in preserving accurate structural details.}
\label{fig:related_sisr}
\end{figure}

In SISR task, transformer-based SISR methods~\cite{zhou2023srformer,chen2023hat,9607618,Hsu_2024_CVPR,Chu_2024_CVPR} achieve better performance than traditional CNN-based methods~\cite{8099502,Lim_2017_CVPR_Workshops,NEURIPS2020_eaae339c,wang2018esrgan, zhang2018image}. 
\paragraph{Swin-transformer-based Method.}
SwinIR~\cite{9607618}, a classic Swin-transformer-based~\cite{vaswani2017attention,liu2021swin} SISR model, performs self-attention within $8 \times 8$ local windows to extract deep features that significantly helps improve reconstructed high-resolution results. DRCT~\cite{Hsu_2024_CVPR} adds dense-connections~\cite{DenseNet2017} inside its residual blocks based on the architecture of SwinIR to mitigate the spatial information vanishing in deep feature extraction modules. HMANet~\cite{Chu_2024_CVPR} aims to extract self-similarity of images, including local similarity between nearby regions and global similarity across distant areas, by adding grid-attention to window-based self-attention mechanism.
\paragraph{GAN-based Method.}    
Although GAN-based~\cite{wang2018esrgan,ledig2017photo,wu2017srpgan,zhang2021designing} methods are able to generate visually richer textures, the adversarial loss tends to encourage the generator to produce plausible-looking details~\cite{ledig2017photo,zhang2018unreasonable}, even if those details contradict the ground-truth. To alleviate this problem, ESRGAN~\cite{wang2018esrgan} introduces pixel-wise loss in its loss function, but structure distortions and hallucinated details still exist demonstrated by \cref{fig:related_sisr}.
\paragraph{Diffusion-based Method.}    
Same situation happens to diffusion-based SISR models~\cite{li2022srdiff,saharia2022image,yue2023resshift,shang2024resdiff,yue2025arbitrary,wang2024sinsr}. Those models require iterative noise injection and denoising when predicting high-resolution images, which tends to alter images structure and details. The diffusion process mechanism could break the alignment across different material maps, making it less suitable for PBR material super-resolution task.

Considering the importance of both pixel-level accuracy and cross-map alignment in PBR material SR task, we adopt Swin-transformer-based architectures as our backbones.

\subsection{PBR Materiel Super-Resolution}
PBR material SR is a challenging problem due to the unique constraints of material maps, requiring strict cross-map consistency, and presenting structural patterns distinct from natural images. These factors have limited extensive research on this topic. Recently, MatUp~\cite{gauthier2024matup} attempts to address PBR material SR by leveraging SOTA SISR models. It generates pseudo ground-truth through SOTA SISR models, which subsequently supervise a multilayer perceptron (MLP) via render loss. However, MatUp suffers from two key limitations when applying to real-world PBR material SR task:
\begin{itemize}
\item MatUp trains a dedicated MLP model for each material at inference time. This process is time-consuming and limits generalization.
\item The simple architecture of the MLP struggles to capture the complexity of PBR material SR task, making outputs quality significantly below the pseudo ground-truth generated by SISR models.
\end{itemize}

PBR-SR~\cite{chen2025pbr}, another PBR material SR method, aims to utilize pre-trained SISR models in a similar way as MatUp to upscale mesh textures. However, it remains a mesh-specific model, making its practical application computationally expensive. Similarly to MatUp, PBR-SR suffers from the inherent limitation that the quality of its outputs cannot surpass that of the high-resolution pseudo ground-truth generated by the pre-trained SISR model.

As opposed to MatUp and PBR-SR, which require training a dedicated model or optimization process for each material, our MUJICA approach offers superior generalization with material-agnostic capabilities, enabling efficient and scalable applications across diverse PBR materials in industrial applications.

\subsection{Multi-Modal Fusion}
Traditionally, multi-modal fusion strategies are classified, based on the stage at which fusion occurs~\cite{fusion_strategy_01, fusion_strategy_02, fusion_strategy_survey, baltruvsaitis2018multimodal,boulahia2021early,guarrasi2025systematic}, as early fusion, intermediate fusion, and late fusion.

Early fusion, also referred as data-level fusion, involves a simple concatenation of different modalities as input to form a shared feature subspace. Besides, early fusion prefers single-stream architecture~\cite{gauthier2024matup, zhang2023superyolo, zhao2020single, zhang2020uc}. For example, SuperYOLO~\cite{zhang2023superyolo} employs pixel-wise concatenation between infrared images and RGB images to enhance super-resolution performance, improving object detection accuracy. MatUp~\cite{gauthier2024matup} also adopts this data-level fusion, however, it does not perform optimally, because PBR materials are rendered according to complex rendering functions like SVBRDF and exhibit intricate interdependence that may not naturally align in the image space. For instance, a wrinkled white paper may have a highly detailed normal map while its basecolor map remains nearly uniform white. Such scenarios require more sophisticated modeling of modal interactions. Simply concatenating them at input level may ignore the difference between different modalities and fail to capture the comprehensive cross-map interaction.

Recent study~\cite{guo2024cmdaf, tsai2019multimodal} demonstrates that feature-level attention-based fusion strategy with two-stream architecture often outperforms early fusion in tasks of loosely coupled or heterogeneous modalities. 

In this work, we employ an intermediate fusion strategy by first performing cross-modal interactions, followed by modality-specific feature extraction to preserve and refine distinct characteristics.

\section{Method}
\label{sec:method}

In this section, $m \in \mathcal{M}$ is defined as different PBR material map such as basecolor, normal, roughness and metallic, where $\mathcal{M}$ denotes a collection of all maps of the PBR material. $n$ is the PBR material map in set $\mathcal{M}\setminus \{m\}$ to provide complementary features. In this paper, each map is considered as an independent modality of the PBR material to better modeling the task.

\begin{figure}[hbt]
    \centering
    \includegraphics[width=0.95\linewidth, trim=15 15 20 15, clip]{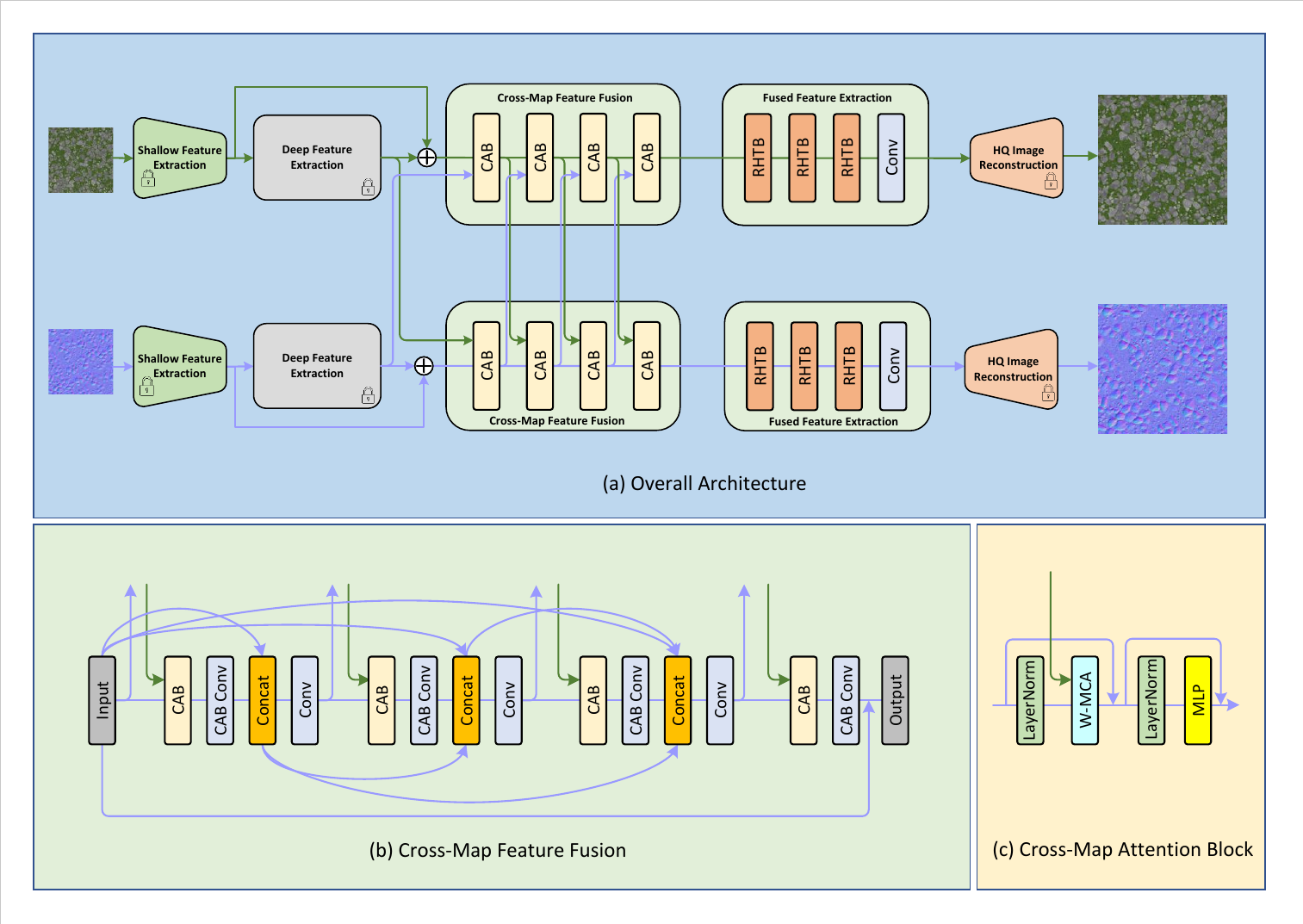} 
    \caption{Architecture overview. (a) Illustration of the interaction between \textcolor[rgb]{0.40,0.56,0.29}{basecolor} and \textcolor[rgb]{0.65,0.66,0.98}{normal} modalities. The lock symbol indicates modules that are frozen during training. (b) and (c) only display of the \textcolor[rgb]{0.65,0.66,0.98}{normal} modality.}
    \label{fig:architecture}
\end{figure}

As shown in \cref{fig:architecture}, our network consists of 5 modules. Among them, Shallow and Deep Feature Extraction, along with the HQ Image Reconstruction modules, are frozen during training. In contrast, Cross-Map Feature Fusion and Fused Feature Extraction modules, called MUJICA, are trainable adapter modules.

\subsection{Overall Architecture}

\paragraph{Shallow and Deep Feature Extraction.}
Given a low-resolution (LR) input $\mathbf{I}_{LR}^m \in \mathbb{R}^{H \times W \times C_{in}}$ ($H$, $W$ and $C_{in}$ are the image height, width and input channel number, respectively) for $m \in \mathcal{M}$, the shallow feature $\mathbf{F}_0^m \in \mathbb{R}^{H \times W \times C}$ is extracted by a shared $3 \times 3$ convolutional layer~\cite{NEURIPS2021_ff1418e8} $\text{Conv}(\cdot)$, where $C$ is the feature channel number.

$\mathbf{F}_0^m$ is then passed through a shared deep Swin-transformer-based backbone $H_{DF}(\cdot)$ to obtain modality-specific deep feature $\mathbf{F}_{DF}^m$. The resulting deep feature captures rich spatial information essential for texture and detail reconstruction.

\paragraph{Cross-Map Feature Fusion.}
To enhance the modality-specific feature $\mathbf{F}_{Fused}^m$ with cross-modality context, we propose the Cross-Map Feature Fusion module, consisting of $L$ stacked Cross-Map Attention Blocks (CABs), denoted as $\{B_l^m\}_{l=1}^L$. At each layer, features of modality $m$ are refined using both self and other modality features. The fusion process is formulated as
\begin{equation*}\label{eq:sr_feature_fusion}
    \mathbf{F}_{Fused}^m = H_{FF}^m(\mathbf{F}_{DF}^m + \mathbf{F}_{0}^m, \mathbf{F}_{DF}^n),
\end{equation*}
where $H_{FF}(\cdot, \cdot)$ denotes the feature fusion module for modality $m$; the external set $\{\mathbf{F}_{DF}^n\}$ provides complementary features from other modalities to enhance the representation of modality $m$.

\paragraph{Fused Feature Extraction.}
After Cross-Map Feature Fusion module, the modality-specific representation is further enhanced by applying a sequence of residual transformer blocks. These blocks are designed to extract features from $\mathbf{F}_{Fused}^m$. For a given modality $m$, the extracted fused feature $\mathbf{F}_{FFE}^m$ is calculated by $H_{FFE}^m(\mathbf{F}_{Fused}^m)$, where $H_{FFE}^m(\cdot)$ denotes the modality-specific Fused Feature Extraction module.

\paragraph{HQ Image Reconstruction.}
The high-quality super-resolution (SR) image $I^m_{SR}$ for modality $m$ is reconstructed from $\mathbf{F}_{FFE}^m$ using a shared reconstruction module $H_{rec}(\cdot)$, where $H_{rec}(\cdot)$ is implemented using pixel shuffle and convolutional layers.

\subsection{Cross-Map Feature Fusion}
The Cross-Map Feature Fusion module comprises $L$ stacked Cross-Map Attention Blocks (CABs) $B_l$, with each performing layer-wise interaction between modality $m$ and $n$. Inspired by DRCT~\cite{Hsu_2024_CVPR} which demonstrates that, compared to residual connections, dense connections better preserve high-frequency features during cross-map feature fusion. Hence, a dense connection strategy is adopted in this module to progressively concatenate intermediate features before passing them to next CAB.

Different from previous multi-modal fusion methods which typically perform cross-modal interaction only at selected stages, our proposed Cross-Map Feature Fusion module emphasizes progressive and layer-wise fusion. Within each CAB, modality features are iteratively refined using complementary context from other modalities, allowing proposed network to gradually align different PBR map features. Let $x_l^m \in \mathbb{R}^{H \times W \times C}$ denote the feature of the given modality $m$ produced by $l$-th CAB, the whole cross-map feature fusion process is formulated as
\begin{equation*}\label{eq:sr_feature_fusion_detail}
\begin{aligned}
x_1^m &= B_1^m\big(
  [\mathbf{F}_{DF}^m + \mathbf{F}_{0}^m], \mathbf{F}_{DF}^n
\big), \\
x_2^m &= B_2^m\big(
  [\mathbf{F}_{DF}^m + \mathbf{F}_{0}^m,\ H_1^m(x_1^m)],\ x_1^n 
\big), \\
&~..., \\
x_L^m &= B_L^m\big(
  [\mathbf{F}_{DF}^m + \mathbf{F}_{0}^m,\ H_1^m(x_1^m),\ \dots,\ H_{L-1}^m(x_{L-1}^m)], \\
&\qquad\quad  x_{L-1}^n
\big),
\end{aligned}
\end{equation*}
where $[\cdot, ..., \cdot]$ denotes the concatenation of features; for $i\in\{1, 2, ..., L-1\}$, $H_i^m(\cdot)$, a transition function to reduce the dimensionality of the previous layers' outputs and prevent feature explosion, is implemented as a $1 \times 1$ convolutional layer along with a LeakyReLU~\cite{xu2015empirical} with negative slope of 0.2. Despite the progressive dense fusion and repeated cross-map attention, our proposed CAB maintains moderate computational complexity, thanks to compact transition functions $H_i^m(\cdot)$. For a given $i\in\{1, 2, ..., L-1\}$, the set $\{x_i^n\}$ provides complementary information from other modalities to enhance the representation of modality $m$ at $(i+1)$-th CAB.

After all CABs, a residual connection mechanism is applied to aggregate $x_L^m$ with the original input $(\mathbf{F}_{DF}^m + \mathbf{F}_{0}^m)$ to calculate the final output $\mathbf{F}_{Fused}^m$ as
\begin{equation*}\label{eq:sr_feature_fusion_residual}
\mathbf{F}_{Fused}^m = \alpha \cdot x_L^m + (\mathbf{F}_{DF}^m + \mathbf{F}_{0}^m),
\end{equation*}
where $\alpha$ is a learnable scaling factor to stabilize the training process.

\subsection{Cross-Map Attention Block}
To balances expressive ability and efficiency, our proposed CABs are implemented using efficient window attention. A Window-based Multi-head Cross-map Attention (W-MCA) mechanism is introduced to enable modality-aware interaction. Suppose input size is $H \times W \times C$, Swin Transformer~\cite{liu2021swin} firstly partitions the input into $\frac{H\times W}{S^2}\times S^2\times C$ non-overlapping $S \times S$ local widows. Given an embedding dimension $d$ and a local window $X_m \in \mathbb{R}^{S^2 \times C}$ of current modality $m$, $\forall i \in \mathcal{M}$, the Query, Key and Value, $Q_i$, $K_m$ and $V_m \in \mathbb{R}^{S^2 \times d}$ matrices are computed as
\begin{equation*}
Q_i = X_i P_{Q_i},~K_m = X_m P_{K_m},~V_m = X_m P_{V_m},
\label{eq:sr_cross_attention_detail}
\end{equation*}
where $X_i \in \mathbb{R}^{S^2 \times C}$ is the local window from all modalities $\mathcal{M}$; $P_{Q_i}$, $P_{K_m}$ and $P_{V_m}\in \mathbb{R}^{C \times d}$ are linear projection matrices shared across different windows. Note that $P_{K_m}$ and $P_{V_m}$ are applied only to the local window input $X_m$ of modality $m$, while $P_{Q_i}$ is specific to each query modality $i$, being applied to its corresponding local window input $X_i$.

For $m \in \mathcal{M}$, the cross-map attention is calculated as
\begin{equation*}
\text{Attn}_m = \sum_{i \in \mathcal{M}} \mathbf{SoftMax}\left( \frac{Q_i (K_m)^T}{\sqrt{d}} + b_i \right) V_m,
\end{equation*}
where $b_i$ is a learnable relative positional encoding.

\subsection{Fused Feature Extraction}
Compared with late fusion, intermediate fusion presents better effectiveness in capturing complex interactions between different modalities in deep learning~\cite{boulahia2021early, guarrasi2025systematic}. Instead of merging features only at the final reconstruction stage, intermediate fusion encourages early and repeated interactions across different modalities, allowing the model to learn richer representation of features.

To leverage this advantage, for modality $m$, a modality-specific Fused Feature Extraction module $H_{FFE}^m(\cdot)$ is attached immediately after the Cross-Map Feature Fusion module. This module extracts refined representations from its own fused feature $F^m_{Fused}$, while preserving multi-modal contextual benefits inherited from the Cross-Map Feature Fusion module. It is worth noting that $H_{FFE}^m(\cdot)$ is a modular and flexible component, aligning with any Deep Feature Extraction architecture employed in Swin-transformer-based SISR models. It can be initialized by those pre-trained SISR models to provide a better starting point and accelerate convergence. This plug-and-play manner allows freely choosing a suitable architecture in terms of computational cost, inference time and reconstruction accuracy. As an example shown in ~\cref{fig:architecture}, the Residual Hybrid Transformer Block (RHTB) from HMANet~\cite{Chu_2024_CVPR} is adopted as base residual transformer block of proposed $H_{FFE}^m(\cdot)$.

\subsection{Loss Function}
A two terms total loss function to supervise the learning of PBR material super-resolution process is defined as
\begin{equation*}\label{eq:total_loss_function}
  \mathcal{L}_{total} = \mathcal{L}_{rec} + \mathcal{L}_{mat},
\end{equation*} 
where $\mathcal{L}_{\text{rec}}$ denotes the reconstruction loss based on rendering appearance and $\mathcal{L}_{\text{mat}}$ refers to the PBR material map loss that directly compares reconstructed SR material maps with ground-truth.

\paragraph{Reconstruction Loss.}
To calculate reconstruction loss $\mathcal{L}_{\text{rec}}$ under consistent illumination, both the SR material maps and ground-truth are rendered under a set $\Omega$ of single point light sources, uniformly sampled from the hemisphere based on the Fibonacci sampling strategy. The $\mathcal{L}_{\text{rec}}$ is then computed between resulting renderings by measuring both pixel-wise and perceptual differences as below
\begin{equation*}\label{eq:reconstruction-loss}
\begin{aligned}
\mathcal{L}_{\text{rec}} =\ 
&\sum_{\omega_i \in \Omega} 
\mathcal{L}_{\text{1c}}\left( 
R^{\omega_i}( \mathcal{M}_{\text{gt}} ),\ 
R^{\omega_i}( \mathcal{M}_{\text{sr}} ) 
\right) \nonumber \\
+\ 
&\sum_{\omega_i \in \Omega} 
\mathcal{L}_{\text{VGG-11}}\left( 
R^{\omega_i}( \mathcal{M}_{\text{gt}} ),\ 
R^{\omega_i}( \mathcal{M}_{\text{sr}} ) 
\right),
\end{aligned}
\end{equation*}
where $R^{\omega_i}\{\cdot\}$ denotes rendering under single point light $\omega_i \in \Omega$; $\mathcal{L}_{\text{1c}}$ is the Charbonnier~\cite{lai2018fast,charbonnier1994two,anagun2019srlibrary} loss for pixel-wise differences and $\mathcal{L}_{\text{VGG-11}}$ is the perceptual loss~\cite{johnson2016perceptual, ledig2017photo, wu2017srpgan, rad2019srobb, zhang2018unreasonable} calculating the differences between features extracted from a pre-trained VGG11 network~\cite{simonyan2014very}; $\mathcal{M}_{\text{gt}}$ and $\mathcal{M}_{\text{sr}}$ are the sets of ground-truth material maps and SR material maps, respectively.

\paragraph{Material Loss.}
 In parallel, material loss $\mathcal{L}_{\text{mat}}$ is introduced to directly compare SR material maps with ground-truth without rendering. It is formulated including both pixel-wise and perceptual loss as
\begin{equation*}\label{eq:material-loss}
\begin{aligned}\mathcal{L}_{\text{mat}} =\ 
&\sum_{m \in \mathcal{M}}
\mathcal{L}_{\text{1c}}\left( 
m_{\text{gt}},\ 
m_{\text{sr}}
\right) \nonumber \\
+\ 
&\sum_{m \in \mathcal{M}}
\mathcal{L}_{\text{VGG-11}}\left( 
m_{\text{gt}},\ 
m_{\text{sr}} 
\right),
\end{aligned}
\end{equation*}
where $m_{\text{gt}}$ and $m_{\text{sr}}$ respectively denote the ground-truth and SR material maps for each map $m$ in $\mathcal{M}$.

The definitions of above Charbonnier and perceptual losses are explained as follows.
\paragraph{Charbonnier Loss.}
The Charbonnier loss~\cite{lai2018fast,charbonnier1994two,anagun2019srlibrary} is a robust and differentiable approximation of the $L1$ loss defined as
\begin{equation*}
  \mathcal{L}_{1c} = \sqrt{\| \mathbf{I}_{\text{gt}} - {\mathbf{I}_{\text{pred}}} \|_2^2 + \epsilon^2},
\end{equation*}
where $\epsilon$ is set to $10^{-3}$; $\mathbf{I}_{\text{gt}}$ and $\mathbf{I}_{\text{pred}}$ denote predicted result and ground-truth label respectively. The Charbonnier loss presents faster convergence than $L2$ loss and better robustness to outliers compared to the $L1$ loss.

\paragraph{Perceptual Loss.}
The perceptual loss has been widely adopted to train super-resolution models~\cite{johnson2016perceptual,ledig2017photo,wu2017srpgan,rad2019srobb,zhang2018unreasonable}. Inspired by recent findings in image super-resolution research~\cite{pihlgren2024systematicperformanceanalysisdeep}, higher weights are assigned to shallower VGG11 layers to better capture high-frequency details such as textures and edges. The perceptual loss is defined as
\begin{equation*}
  \mathcal{L}_{\text{VGG-11}} = \sum_{i \in L} 
  \lambda_i \cdot \mathcal{L}_{\text{1c}}\left( 
    \phi_i\left(\mathbf{I}_{\text{gt}}\right),\ 
    \phi_i\left(\mathbf{I}_{\text{pred}}\right)
  \right),
\end{equation*}
where $L$ denotes the set of selected convolutional layers in the pre-trained VGG11 network; $\phi_i(\cdot)$ denotes feature maps extracted from the $i$-th layer in $L$, and $\lambda_i$ denotes the weight assigned to the $i$-th layer.

\section{Experiments}
\label{sec:experiments}
\subsection{Experimental Setup}
In order to validate the performance of our method, we perform extensive experiments to compare it against existing SISR baselines using two different $2048 \times 2048$ resolution datasets, namely MatSynth~\cite{vecchio2024matsynth} with 5,700 mixed quality assets for pre-training and our in-house dataset with 1,129 high quality and rich detailed assets for fine-tuning. LR versions of GT PBR materials are obtained by applying a bicubic down-sampling method with scale factors $0.5$ and $0.25$. 

We randomly crop images into $64\times64$ patches for training with a batch size of 16 on one NVIDIA H100 (80GB). The number of CAB is set to 4 with 6 attention heads, and the window size of W-MCA is set to $8\times8$. Fused Feature Extraction module is set to 3 depths with 6 attention heads in each. During pre-training based on MatSynth, the number of epochs is set to 170, and the learning rate is initialized to $1\times 10^{-4}$ and halved at $35\%$, $60\%$, $75\%$ and $90\%$. During fine-tuning based on our in-house dataset, the number of epochs is set to 500 with a batch size of 16, and the initial learning rate is set to $4\times 10^{-5}$ and halved at $35\%$, $60\%$, $75\%$ and $90\%$. The model is optimized using the Lion~\cite{NEURIPS2023_9a39b492} optimizer. 

We employ the same renderer for both training and evaluation, implementing the Cook-Torrance BRDF model~\cite{cook1982reflectance} with:
\begin{itemize}
    \item Trowbridge-Reitz GGX normal distribution~\cite{trowbridge1975average},
    \item Schlick-GGX geometry term~\cite{karis2013real},
    \item Schlick's Fresnel approximation~\cite{schlick1994inexpensive}.
\end{itemize}
Unless otherwise specified, in order to ensure fairness and objectivity, the maps not involved in the SR process are replaced with the GT version during the rendering process.

\subsection{Quantitative Comparison}
\Cref{tab:sr_comparison} reports the quantitative comparison between MUJICAs and SOTA SISR methods on both MatSynth and our in-house dataset. Evaluation is performed in terms of PSNR(dB), SSIM and LPIPS~\cite{zhang2018unreasonable}. PSNR and SSIM assess pixel-level fidelity and structural similarity, while LPIPS serves as a perceptual metric aligned with human visual judgment. Those metrics provide a comprehensive and balanced assessment of both reconstruction accuracy and perceptual quality.

MUJICAs consistently achieves superior performance across all metrics and datasets. Notably, HMANet-MUJICA consistently outperforms SwinIR-MUJICA, reflecting the relative capability of their SISR backbones. For $\times 2$ SR task, MUJICAs outperform their corresponding SISR backbones with gains up to \textbf{1.15dB} in PSNR, \textbf{0.0069} in SSIM, and a reduction of \textbf{0.036} in LPIPS on renderings across datasets. For $\times 4$ SR task, MUJICAs improves metrics up to \textbf{0.76dB} in PSNR, \textbf{0.0070} in SSIM, and \textbf{0.0695} in LPIPS. 
These metrics demonstrate that MUJICA consistently outperforms existing SISR models.

\begin{table*}
\centering
\scalebox{0.52}{
\begin{tabular}{|l|c|
    ccc|ccc|ccc|  
    ccc|ccc|ccc|} 
\hline
\multirow{3}{*}{\textbf{Method}} & \multirow{3}{*}{\textbf{Scale}}
  & \multicolumn{9}{c|}{\textbf{MatSynth~\shortcite{vecchio2024matsynth}}}
  & \multicolumn{9}{c|}{\textbf{In-house Dataset}} \\
\cline{3-20}
& & \multicolumn{3}{c|}{Rendering} & \multicolumn{3}{c|}{Basecolor} & \multicolumn{3}{c|}{Normal}
  & \multicolumn{3}{c|}{Rendering} & \multicolumn{3}{c|}{Basecolor} & \multicolumn{3}{c|}{Normal} \\
\cline{3-20}
& & PSNR$\uparrow$ & SSIM$\uparrow$ & LPIPS$\downarrow$ & PSNR$\uparrow$ & SSIM$\uparrow$ & LPIPS$\downarrow$ & PSNR$\uparrow$ & SSIM$\uparrow$ & LPIPS$\downarrow$
  & PSNR$\uparrow$ & SSIM$\uparrow$ & LPIPS$\downarrow$ & PSNR$\uparrow$ & SSIM$\uparrow$ & LPIPS$\downarrow$ & PSNR$\uparrow$ & SSIM$\uparrow$ & LPIPS$\downarrow$ \\
\hline
HAT~\shortcite{chen2023hat} & $\times2$
  & 37.51 & 0.9359 & 0.0842 & \colorbox[rgb]{1.00,1.00,0.66}{43.75} & 0.9624 & 0.0668 & 38.20 & 0.9404 & 0.0391 
  & 36.27 & 0.8963 & 0.1257 & 36.44 & \colorbox[rgb]{1.00,1.00,0.66}{0.8843} & 0.1661 & 39.34 & 0.9425 & 0.0317 \\
SRFormer~\shortcite{zhou2023srformer} & $\times2$
  & 37.29 & 0.9355 & 0.0859 & 43.72 & \colorbox[rgb]{1.00,1.00,0.66}{0.9625} & 0.0674 & 37.81 & 0.9392 & 0.0402
  & 36.03 & 0.8956 & 0.1286 & 36.29 & 0.8841 & 0.1688 & 38.77 & 0.9416 & 0.0322 \\
SwinIR~\shortcite{9607618} & $\times2$
  & 37.17 & 0.9332 & 0.0885 & 43.62 & 0.9621 & 0.0678 & 37.66 & 0.9359 & 0.0437
  & 36.02 & 0.8939 & 0.1299 & 36.31 & 0.8837 & 0.1693 & 38.68 & 0.9386 & 0.0344 \\
HMANet~\shortcite{Chu_2024_CVPR} & $\times2$
  & \colorbox[rgb]{1.00,1.00,0.66}{37.70} & \colorbox[rgb]{1.00,0.84,0.67}{0.9375} & \colorbox[rgb]{1.00,1.00,0.66}{0.0806} & \colorbox[rgb]{1.00,0.84,0.67}{43.86} & \colorbox[rgb]{1.00,0.84,0.67}{0.9629} & \colorbox[rgb]{1.00,1.00,0.66}{0.0652} & \colorbox[rgb]{1.00,1.00,0.66}{38.54} & \colorbox[rgb]{1.00,1.00,0.66}{0.9430} & \colorbox[rgb]{1.00,1.00,0.66}{0.0356}
  & \colorbox[rgb]{1.00,1.00,0.66}{36.28} & \colorbox[rgb]{1.00,1.00,0.66}{0.8986} & \colorbox[rgb]{1.00,1.00,0.66}{0.1210} & \colorbox[rgb]{1.00,1.00,0.66}{36.52} & \colorbox[rgb]{1.00,0.84,0.67}{0.8856} & \colorbox[rgb]{1.00,1.00,0.66}{0.1636} & \colorbox[rgb]{1.00,1.00,0.66}{39.40} & \colorbox[rgb]{1.00,1.00,0.66}{0.9451} & \colorbox[rgb]{1.00,1.00,0.66}{0.0288} \\
\textcolor[rgb]{0.42,0.35,0.80}{SwinIR-MUJICA} & $\times2$
  & \colorbox[rgb]{1.00,0.84,0.67}{37.77} & \colorbox[rgb]{1.00,1.00,0.66}{0.9362} & \colorbox[rgb]{1.00,0.84,0.67}{0.0672} & 43.63 & 0.9612 & \colorbox[rgb]{1.00,0.84,0.67}{0.0561} & \colorbox[rgb]{1.00,0.84,0.67}{39.38} & \colorbox[rgb]{1.00,0.84,0.67}{0.9491} & \colorbox[rgb]{1.00,0.68,0.69}{0.0230}
  & \colorbox[rgb]{1.00,0.84,0.67}{37.17} & \colorbox[rgb]{1.00,0.84,0.67}{0.9008} & \colorbox[rgb]{1.00,0.84,0.67}{0.0980} & \colorbox[rgb]{1.00,0.84,0.67}{36.94} & 0.8842 & \colorbox[rgb]{1.00,0.84,0.67}{0.1420} & \colorbox[rgb]{1.00,0.84,0.67}{43.26} & \colorbox[rgb]{1.00,0.84,0.67}{0.9558} & \colorbox[rgb]{1.00,0.84,0.67}{0.0135} \\
\textcolor[rgb]{0.42,0.35,0.80}{HMANet-MUJICA} & $\times2$
  & \colorbox[rgb]{1.00,0.68,0.69}{38.21} & \colorbox[rgb]{1.00,0.68,0.69}{0.9389} & \colorbox[rgb]{1.00,0.68,0.69}{0.0613} & \colorbox[rgb]{1.00,0.68,0.69}{44.26} & \colorbox[rgb]{1.00,0.68,0.69}{0.9630} & \colorbox[rgb]{1.00,0.68,0.69}{0.0486} & \colorbox[rgb]{1.00,0.68,0.69}{39.46} & \colorbox[rgb]{1.00,0.68,0.69}{0.9494} & \colorbox[rgb]{1.00,0.84,0.67}{0.0239}
  & \colorbox[rgb]{1.00,0.68,0.69}{37.38} & \colorbox[rgb]{1.00,0.68,0.69}{0.9047} & \colorbox[rgb]{1.00,0.68,0.69}{0.0850} & \colorbox[rgb]{1.00,0.68,0.69}{37.15} & \colorbox[rgb]{1.00,0.68,0.69}{0.8886} & \colorbox[rgb]{1.00,0.68,0.69}{0.1185} & \colorbox[rgb]{1.00,0.68,0.69}{43.90} & \colorbox[rgb]{1.00,0.68,0.69}{0.9575} & \colorbox[rgb]{1.00,0.68,0.69}{0.0127} \\
\hline
ESRGAN~\shortcite{wang2018esrgan} & $\times4$
  & 31.98 & 0.7946 & 0.2923 & \colorbox[rgb]{1.00,0.84,0.67}{38.93} & 0.8719 & 0.2301 & 32.57 & 0.7905 & 0.2108
  & 32.78 & 0.7661 & 0.3302 & 33.57 & 0.7668 & 0.4203 & 35.19 & 0.8298 & 0.1710 \\
SinSR~\shortcite{wang2024sinsr} & $\times4$
  & 24.06 & 0.4941 & 0.4436 & 28.03 & 0.6181 & 0.3963 & 24.58 & 0.5635 & 0.3022
  & 24.53 & 0.4716 & 0.4202 & 25.65 & 0.4801 & 0.4380 & 25.72 & 0.6284 & 0.2801 \\
InvSR~\shortcite{yue2025arbitrary} & $\times4$
  & 23.75 & 0.5435 & 0.3948 & 26.73 & 0.6592 & 0.3707 & 25.17 & 0.6044 & 0.3842
  & 24.05 & 0.5279 & 0.3817 & 24.20 & 0.5209 & 0.4197 & 26.85 & 0.6776 & 0.3082 \\
ResShift~\shortcite{yue2023resshift} & $\times4$
  & 27.94 & 0.6672 & 0.2530 & 33.32 & 0.7633 & \colorbox[rgb]{1.00,0.84,0.67}{0.1877} & 28.37 & 0.7154 & \colorbox[rgb]{1.00,0.84,0.67}{0.1512}
  & 28.70 & 0.6721 & \colorbox[rgb]{1.00,0.68,0.69}{0.2031} & 30.10 & 0.6615 & \colorbox[rgb]{1.00,0.68,0.69}{0.2113} & 29.32 & 0.7757 & 0.1324 \\
HAT~\shortcite{chen2023hat} & $\times4$
  & 32.13 & 0.8040 & 0.2719 & 38.58 & 0.8739 & 0.2235 & 32.79 & 0.8096 & 0.1870
  & 32.82 & 0.7751 & 0.3167 & 33.41 & 0.7703 & 0.4089 & 34.81 & 0.8442 & 0.1547 \\
SRFormer~\shortcite{zhou2023srformer} & $\times4$
  & 32.05 & 0.8029 & 0.2786 & 38.66 & \colorbox[rgb]{1.00,1.00,0.66}{0.8742} & 0.2242 & 32.62 & 0.8062 & 0.1961
  & 32.61 & 0.7735 & 0.3219 & 33.43 & 0.7694 & 0.4150 & 34.03 & 0.8421 & 0.1600 \\
SwinIR~\shortcite{9607618} & $\times4$
  & 31.99 & 0.7995 & 0.2795 & 38.52 & 0.8723 & 0.2236 & 32.59 & 0.8026 & 0.1976
  & 32.68 & 0.7715 & 0.3229 & 33.37 & 0.7684 & 0.4126 & 34.23 & 0.8389 & 0.1591 \\
DRCT~\shortcite{Hsu_2024_CVPR} & $\times4$
  & 31.97 & 0.7985 & 0.2878 & 38.40 & 0.8720 & 0.2264 & 32.60 & 0.7992 & 0.2080
  & 32.72 & 0.7693 & 0.3282 & 33.38 & 0.7672 & 0.4157 & 34.51 & 0.8360 & 0.1708 \\
HMANet~\shortcite{Chu_2024_CVPR} & $\times4$
  & \colorbox[rgb]{1.00,0.84,0.67}{32.37} & \colorbox[rgb]{1.00,0.84,0.67}{0.8093} & 0.2702 & \colorbox[rgb]{1.00,1.00,0.66}{38.73} & \colorbox[rgb]{1.00,0.84,0.67}{0.8764} & 0.2203 & 33.20 & 0.8167 & 0.1869
  & 32.85 & \colorbox[rgb]{1.00,0.84,0.67}{0.7789} & 0.3139 & 33.46 & 0.7733 & 0.4030 & 34.79 & 0.8480 & 0.1554 \\
\textcolor[rgb]{0.42,0.35,0.80}{SwinIR-MUJICA} & $\times4$
  & \colorbox[rgb]{1.00,1.00,0.66}{32.30} & \colorbox[rgb]{1.00,1.00,0.66}{0.8041} & \colorbox[rgb]{1.00,0.84,0.67}{0.2398} & 38.42 & 0.8678 & \colorbox[rgb]{1.00,1.00,0.66}{0.2043} & \colorbox[rgb]{1.00,0.84,0.67}{33.88} & \colorbox[rgb]{1.00,0.84,0.67}{0.8293} & \colorbox[rgb]{1.00,1.00,0.66}{0.1581}
  & \colorbox[rgb]{1.00,0.84,0.67}{33.20} & \colorbox[rgb]{1.00,1.00,0.66}{0.7783} & 0.2698 & \colorbox[rgb]{1.00,0.84,0.67}{33.61} & \colorbox[rgb]{1.00,0.84,0.67}{0.7737} & 0.3387 & \colorbox[rgb]{1.00,1.00,0.66}{36.68} & \colorbox[rgb]{1.00,0.84,0.67}{0.8580} & \colorbox[rgb]{1.00,0.84,0.67}{0.1197} \\
\textcolor[rgb]{0.42,0.35,0.80}{DRCT-MUJICA} & $\times4$
  & 32.23 & 0.8022 & \colorbox[rgb]{1.00,1.00,0.66}{0.2452} & 38.44 & 0.8662 & 0.2118 & \colorbox[rgb]{1.00,1.00,0.66}{33.63} & \colorbox[rgb]{1.00,1.00,0.66}{0.8263} & 0.1613
  & \colorbox[rgb]{1.00,1.00,0.66}{33.11} & 0.7762 & \colorbox[rgb]{1.00,1.00,0.66}{0.2649} & \colorbox[rgb]{1.00,1.00,0.66}{33.58} & \colorbox[rgb]{1.00,1.00,0.66}{0.7734} & \colorbox[rgb]{1.00,1.00,0.66}{0.3317} & \colorbox[rgb]{1.00,0.84,0.67}{36.84} & \colorbox[rgb]{1.00,1.00,0.66}{0.8579} & \colorbox[rgb]{1.00,1.00,0.66}{0.1214} \\
\textcolor[rgb]{0.42,0.35,0.80}{HMANet-MUJICA} & $\times4$
  & \colorbox[rgb]{1.00,0.68,0.69}{32.73} & \colorbox[rgb]{1.00,0.68,0.69}{0.8148} & \colorbox[rgb]{1.00,0.68,0.69}{0.2162} & \colorbox[rgb]{1.00,0.68,0.69}{39.24} & \colorbox[rgb]{1.00,0.68,0.69}{0.8773} & \colorbox[rgb]{1.00,0.68,0.69}{0.1748} & \colorbox[rgb]{1.00,0.68,0.69}{34.18} & \colorbox[rgb]{1.00,0.68,0.69}{0.8334} & \colorbox[rgb]{1.00,0.68,0.69}{0.1459}
  & \colorbox[rgb]{1.00,0.68,0.69}{33.61} & \colorbox[rgb]{1.00,0.68,0.69}{0.7859} & \colorbox[rgb]{1.00,0.84,0.67}{0.2444} & \colorbox[rgb]{1.00,0.68,0.69}{34.42} & \colorbox[rgb]{1.00,0.68,0.69}{0.7828} & \colorbox[rgb]{1.00,0.84,0.67}{0.3039} & \colorbox[rgb]{1.00,0.68,0.69}{37.16} & \colorbox[rgb]{1.00,0.68,0.69}{0.8622} & \colorbox[rgb]{1.00,0.68,0.69}{0.1051} \\
\hline
\end{tabular}}
\caption{Quantitative comparison on $\times2$ and $\times4$ PBR material SR task. We highlight the \colorbox[rgb]{1.00,0.68,0.69}{best}, \colorbox[rgb]{1.00,0.84,0.67}{second-best} and \colorbox[rgb]{1.00,1.00,0.66}{third-best} for each metric. \textcolor[rgb]{0.42,0.35,0.80}{Methods} with the suffix "\textcolor[rgb]{0.42,0.35,0.80}{-MUJICA}" indicate the application of MUJICA.}
\label{tab:sr_comparison}
\end{table*}

\subsection{Visual Comparison}
\begin{figure}[htb]
\centering
    \includegraphics[width=1\linewidth]{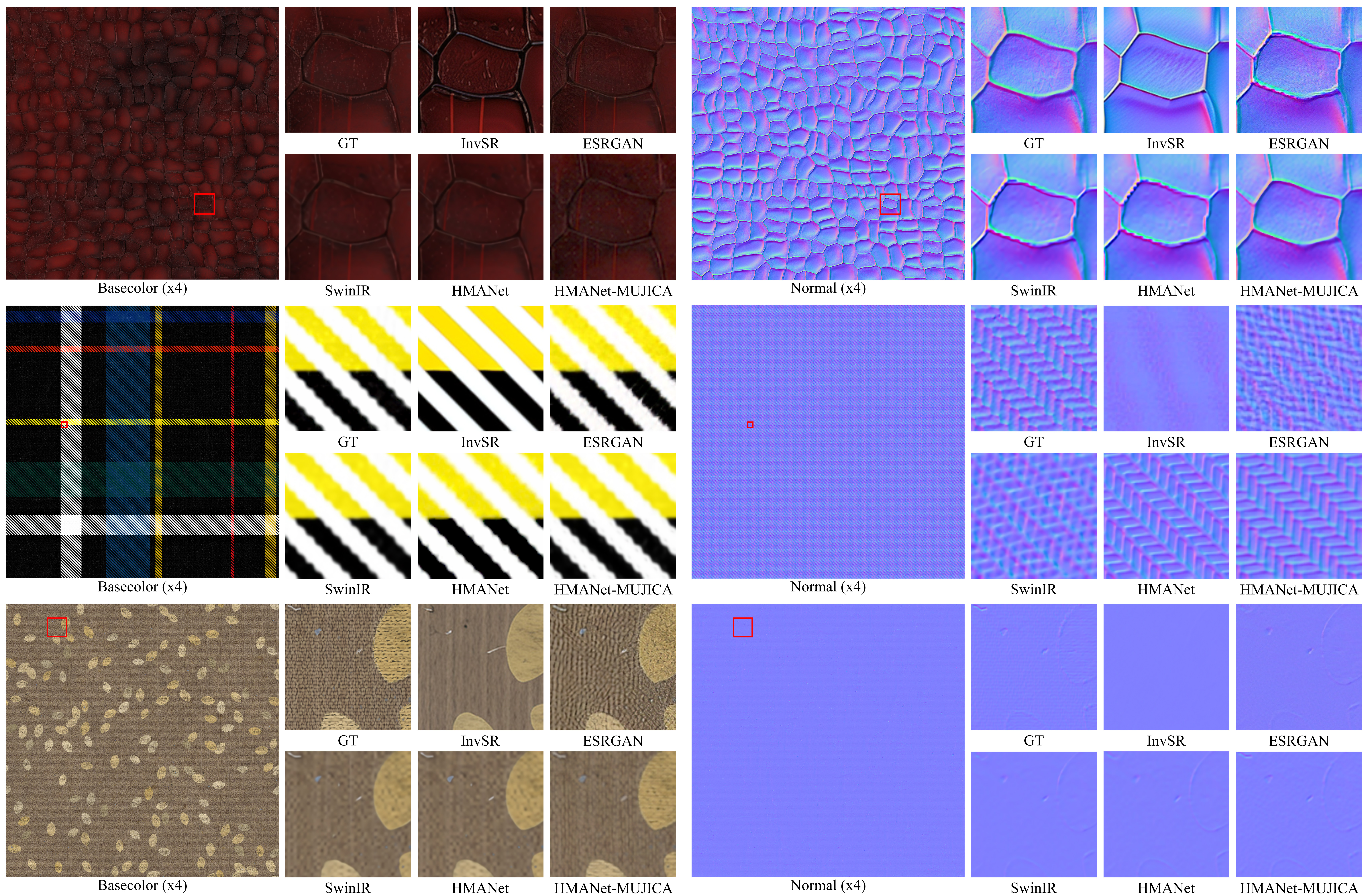}
\caption{Material Map Visual Comparison. MUJICA outperforms other SISR methods in restoring more accurate and more detailed results.}
\label{fig:main_map_visual_comparison}
\end{figure}

\paragraph{Material Map Visual Comparison.}
As presented in \cref{fig:main_map_visual_comparison} and supp.mat fig.1, MUJICA reconstructs finer and more accurate details than its SISR backbone, demonstrating its superior restoration capability. This performance emphasizes the effectiveness of cross-map interactions in enhancing details reconstruction.

\begin{figure}[htb]
\centering
    \includegraphics[width=1\linewidth]{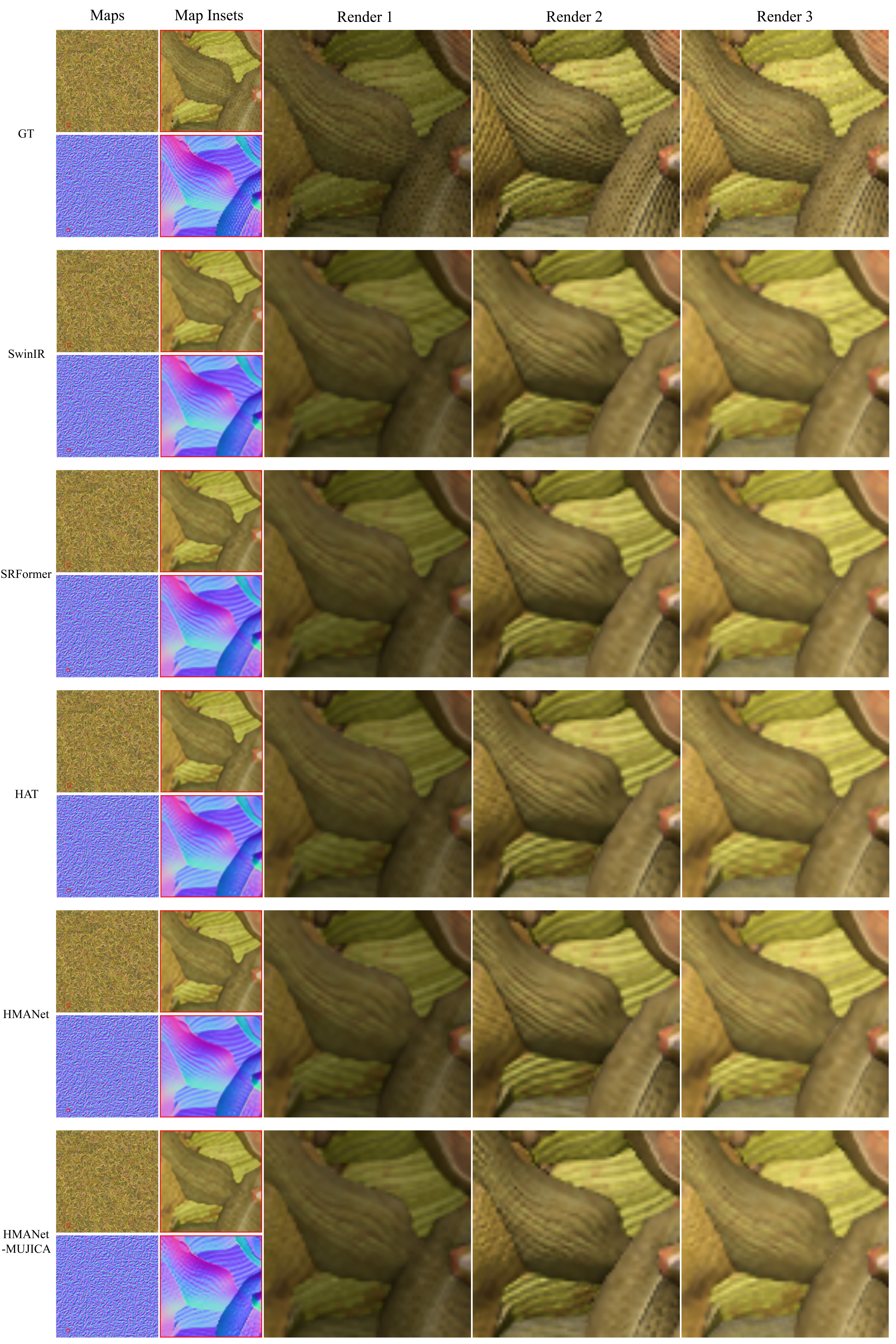}
\caption{Consistency Visual Comparison. Only MUJICA preserves consistency under varying lighting conditions, while others present visible inconsistencies.}
\label{fig:main_render_visual_comparison}
\end{figure}

\paragraph{Consistency Visual Comparison.}
To evaluate rendering consistency, we compare MUJICA with SISR baselines under varying lighting conditions. We use three single point lights selected from Fibonacci-sampled light sets. As demonstrated in \cref{fig:main_render_visual_comparison}, supp.mat fig.2 and supp.mat fig.3, existing SISR models show visible inconsistencies under varying lighting conditions, revealing limitations in their per-map processing manner. In contrast, MUJICA maintains remarkable consistency while restoring better details.

\subsection{Ablation Study}
We choose SwinIR-MUJICA for ablation study based on MatSynth for both training and testing. The SR task is limited to $\times2$. The number of epochs is set to 120 with with a batch size of 8 and evaluation is based on a directional light. 

\paragraph{Impact of Feature Fusion Method.}
We compare our proposed W-MCA with a baseline feature fusion method where the feature maps from different material maps are concatenated and then compressed via a convolution layer. As shown in \cref{tab:ablation_feature_fusion}, replacing the concatenation operation with W-MCA leads to consistent improvements across all metrics, indicating the effectiveness of the attention-based fusion mechanism.

\begin{table}[H]
\scriptsize
\centering
\scalebox{0.7}{
\renewcommand{\arraystretch}{1.1}
\setlength{\tabcolsep}{2.5pt}
\begin{tabular}{|l|>{\hspace{1pt}}c<{\hspace{1pt}}>{\hspace{1pt}}c<{\hspace{1pt}}>{\hspace{1pt}}c<{\hspace{1pt}}|>{\hspace{1pt}}c<{\hspace{1pt}}>{\hspace{1pt}}c<{\hspace{1pt}}>{\hspace{1pt}}c<{\hspace{1pt}}|>{\hspace{1pt}}c<{\hspace{1pt}}>{\hspace{1pt}}c<{\hspace{1pt}}>{\hspace{1pt}}c<{\hspace{1pt}}|}
\hline
\multirow{2}{*}{Methods} &
\multicolumn{3}{c|}{\textbf{Rendering}} & 
\multicolumn{3}{c|}{\textbf{Basecolor}} & 
\multicolumn{3}{c|}{\textbf{Normal}} \\
\cline{2-10}
& \makebox[3.5em]{PSNR$\uparrow$} & \makebox[3.5em]{SSIM$\uparrow$} & \makebox[3.5em]{LPIPS$\downarrow$} 
& \makebox[3.5em]{PSNR$\uparrow$} & \makebox[3.5em]{SSIM$\uparrow$} & \makebox[3.5em]{LPIPS$\downarrow$} 
& \makebox[3.5em]{PSNR$\uparrow$} & \makebox[3.5em]{SSIM$\uparrow$} & \makebox[3.5em]{LPIPS$\downarrow$} \\
\hline
Concat + Conv & 35.59 & 0.9205 & 0.0987 & 40.30 & 0.9473 & 0.0949 & 39.27 & 0.9472 & 0.0260 \\
W-MCA & 37.94 & 0.9374 & 0.0723 & 43.96 & 0.9614 & 0.0646 & 39.74 & 0.9492 & 0.0240 \\
\hline
\end{tabular}}
\vspace{-2mm}
\caption{Ablation study of different feature fusion methods.}
\label{tab:ablation_feature_fusion}
\vspace{-0.3cm}
\end{table}

\paragraph{Impact of CAB Count.}
Metrics of the \cref{tab:ablation_CAB_number} demonstrate that the number of CABs is positively correlated with model performance. Employing 4 CABs provides a good trade-off between performance and computational cost.

\begin{table}[H]
\scriptsize
\centering
\scalebox{0.75}{
\renewcommand{\arraystretch}{1.1}
\setlength{\tabcolsep}{2.5pt}
\begin{tabular}{|l|>{\hspace{1pt}}c<{\hspace{1pt}}>{\hspace{1pt}}c<{\hspace{1pt}}>{\hspace{1pt}}c<{\hspace{1pt}}|>{\hspace{1pt}}c<{\hspace{1pt}}>{\hspace{1pt}}c<{\hspace{1pt}}>{\hspace{1pt}}c<{\hspace{1pt}}|>{\hspace{1pt}}c<{\hspace{1pt}}>{\hspace{1pt}}c<{\hspace{1pt}}>{\hspace{1pt}}c<{\hspace{1pt}}|}
\hline
\multirow{2}{*}{CABs} &
\multicolumn{3}{c|}{\textbf{Rendering}} & 
\multicolumn{3}{c|}{\textbf{Basecolor}} & 
\multicolumn{3}{c|}{\textbf{Normal}} \\
\cline{2-10}
& \makebox[3.5em]{PSNR$\uparrow$} & \makebox[3.5em]{SSIM$\uparrow$} & \makebox[3.5em]{LPIPS$\downarrow$} 
& \makebox[3.5em]{PSNR$\uparrow$} & \makebox[3.5em]{SSIM$\uparrow$} & \makebox[3.5em]{LPIPS$\downarrow$} 
& \makebox[3.5em]{PSNR$\uparrow$} & \makebox[3.5em]{SSIM$\uparrow$} & \makebox[3.5em]{LPIPS$\downarrow$} \\
\hline
2 & 37.84 & 0.9364 & 0.0728 & 43.77 & 0.9611 & 0.0659 & 39.64 & 0.9489 & 0.0241 \\
4 & 37.94 & 0.9374 & 0.0723 & 43.96 & 0.9614 & 0.0646 & 39.74 & 0.9492 & 0.0240 \\
6 & 38.07 & 0.9380 & 0.0719 & 44.08 & 0.9614 & 0.0644 & 39.73 & 0.9496 & 0.0248 \\
\hline
\end{tabular}}
\vspace{-2mm}
\caption{Ablation study of different number of CABs. }
\label{tab:ablation_CAB_number}
\vspace{-0.3cm}
\end{table}

\paragraph{Impact of Connection Methods in CAB.}
\Cref{tab:ablation_connection} displays metrics of 3 different connection methods in CAB, which are No Residual Connection (NRC), Residual Connection (RC) and Dense Connection (DC). As shown in \cref{tab:ablation_connection}, RC improves the performance over NRC, while DC further enhances results compared to RC alone.

\begin{table}[H]
\scriptsize
\centering
\scalebox{0.73}{
\renewcommand{\arraystretch}{1.1}
\setlength{\tabcolsep}{2.5pt}
\begin{tabular}{|l|>{\hspace{1pt}}c<{\hspace{1pt}}>{\hspace{1pt}}c<{\hspace{1pt}}>{\hspace{1pt}}c<{\hspace{1pt}}|>{\hspace{1pt}}c<{\hspace{1pt}}>{\hspace{1pt}}c<{\hspace{1pt}}>{\hspace{1pt}}c<{\hspace{1pt}}|>{\hspace{1pt}}c<{\hspace{1pt}}>{\hspace{1pt}}c<{\hspace{1pt}}>{\hspace{1pt}}c<{\hspace{1pt}}|}
\hline
\multirow{2}{*}{Methods} &
\multicolumn{3}{c|}{\textbf{Rendering}} & 
\multicolumn{3}{c|}{\textbf{Basecolor}} & 
\multicolumn{3}{c|}{\textbf{Normal}} \\
\cline{2-10}
& \makebox[3.5em]{PSNR$\uparrow$} & \makebox[3.5em]{SSIM$\uparrow$} & \makebox[3.5em]{LPIPS$\downarrow$} 
& \makebox[3.5em]{PSNR$\uparrow$} & \makebox[3.5em]{SSIM$\uparrow$} & \makebox[3.5em]{LPIPS$\downarrow$} 
& \makebox[3.5em]{PSNR$\uparrow$} & \makebox[3.5em]{SSIM$\uparrow$} & \makebox[3.5em]{LPIPS$\downarrow$} \\
\hline
NRC & 15.73 & 0.4795 & 0.8796 & 17.26 & 0.4456 & 0.7711 & 24.17 & 0.6054 & 0.6247 \\
RC & 37.79 & 0.9364 & 0.0727 & 43.81 & 0.9607 & 0.0653 & 39.69 & 0.9486 & 0.0247 \\
DC & 37.94 & 0.9374 & 0.0723 & 43.96 & 0.9614 & 0.0646 & 39.74 & 0.9492 & 0.0240 \\
\hline
\end{tabular}}
\vspace{-2mm}
\caption{Ablation study of different connection methods.}
\label{tab:ablation_connection}
\vspace{-0.3cm}
\end{table}

\paragraph{Impact of Fused Maps.}
\Cref{tab:ablation_fused_maps} presents metrics of different map fusion combinations. Two specific combinations are investigated as below
\begin{itemize}
    \item[(a)] Basecolor + Normal,
    \item[(b)] Basecolor + Normal + Roughness.
\end{itemize}
In combination (a), the roughness map is upscaled independently using the pre-trained SwinIR model. As demonstrated in \cref{tab:ablation_fused_maps}, (b) presents a better performance than (a) thanks to the complementary information from the roughness map, confirming the effectiveness of cross-map fusion. Nevertheless, (a) is employed for evaluation experiments because of its reduced computational overhead and lower resource demands.

\begin{table}[H]
\scriptsize
\centering
\scalebox{0.68}{
\renewcommand{\arraystretch}{1.1}
\setlength{\tabcolsep}{2.5pt}
\begin{tabular}{|l|>{\hspace{1pt}}c<{\hspace{1pt}}>{\hspace{1pt}}c<{\hspace{1pt}}>{\hspace{1pt}}c<{\hspace{1pt}}|>{\hspace{1pt}}c<{\hspace{1pt}}>{\hspace{1pt}}c<{\hspace{1pt}}>{\hspace{1pt}}c<{\hspace{1pt}}|>{\hspace{1pt}}c<{\hspace{1pt}}>{\hspace{1pt}}c<{\hspace{1pt}}>{\hspace{1pt}}c<{\hspace{1pt}}|>{\hspace{1pt}}c<{\hspace{1pt}}>{\hspace{1pt}}c<{\hspace{1pt}}>{\hspace{1pt}}c<{\hspace{1pt}}|}
\hline
\multirow{2}{*}{Methods} &
\multicolumn{3}{c|}{\textbf{Rendering}} & 
\multicolumn{3}{c|}{\textbf{Basecolor}} & 
\multicolumn{3}{c|}{\textbf{Normal}} &
\multicolumn{3}{c|}{\textbf{Roughness}} \\
\cline{2-13}
& \makebox[2.5em]{PSNR$\uparrow$} & \makebox[2.5em]{SSIM$\uparrow$} & \makebox[2.5em]{LPIPS$\downarrow$}
& \makebox[2.5em]{PSNR$\uparrow$} & \makebox[2.5em]{SSIM$\uparrow$} & \makebox[2.5em]{LPIPS$\downarrow$}
& \makebox[2.5em]{PSNR$\uparrow$} & \makebox[2.5em]{SSIM$\uparrow$} & \makebox[2.5em]{LPIPS$\downarrow$}
& \makebox[2.5em]{PSNR$\uparrow$} & \makebox[2.5em]{SSIM$\uparrow$} & \makebox[2.5em]{LPIPS$\downarrow$} \\
\hline
(a) & 37.85 & 0.9366 & 0.0740 & 43.96 & 0.9614 & 0.0646 & 39.74 & 0.9492 & 0.0240 & 34.64 & 0.8535 & 0.1877 \\
(b) & 37.91 & 0.9371 & 0.0733 & 43.99 & 0.9617 & 0.0637 & 39.81 & 0.9498 & 0.0232 & 34.91 & 0.8551 & 0.1556 \\
\hline
\end{tabular}}
\vspace{-2mm}
\caption{Ablation study of different map fusion combinations.}
\label{tab:ablation_fused_maps}
\vspace{-0.3cm}
\end{table}
The metallic map, in contrast, typically appears as a constant (often black) image and lacks discriminative patterns. Its distribution substantially diverges from other maps, making it less suitable for cross-map fusion. Consequently, it is excluded from our maps fusion combinations.

\section{Conclusion}
\label{sec:conclusion}
In this work, we propose \textbf{M}ulti-modal \textbf{U}pscaling \textbf{J}oint \textbf{I}nference via \textbf{C}ross-map \textbf{A}ttention (\textbf{MUJICA}), a flexible adapter that reforms pre-trained Swin-transformer-based SISR models for PBR material super-resolution. MUJICA is seamlessly attached after the pre-trained SISR backbone, which remains entirely frozen. It leverages cross-map attention to fuse features while preserving remarkable reconstruction ability of the pre-trained SISR model. Applied to SISR models such as SwinIR, DRCT and HMANet, MUJICA improves PSNR, SSIM, and LPIPS scores while preserving cross-map consistency. Experiments demonstrate that MUJICA enables efficient training even with limited resources and delivers state-of-the-art performance on PBR material datasets.

\section{Limitations and Future Work}
\label{sec:future_work}
As demonstrated in \cref{tab:ablation_fused_maps}, incorporating the roughness map into cross-map fusion enhances overall performance at the cost of higher computational overhead and resource requirements. One possible way is to adopt distributed training for MUJICA including the roughness map in cross-map fusion, which could yield further improvements. Additionally, the \reflectbox{F}LIP loss~\cite{Andersson2021a}, which is theoretically better aligned with the PBR material SR, could replace the VGG-based perceptual loss to enhance performance. Finally, the core idea of cross-map fusion could be extended to other domains, such as intrinsic image decomposition, image delighting and PBR material decomposition.

\newpage
\bibliography{aaai2026}

\begin{thebibliography}{62}
\providecommand{\natexlab}[1]{#1}

\bibitem[{Anagun, Isik, and Seke(2019)}]{anagun2019srlibrary}
Anagun, Y.; Isik, S.; and Seke, E. 2019.
\newblock SRLibrary: Comparing different loss functions for super-resolution over various convolutional architectures.
\newblock \emph{Journal of Visual Communication and Image Representation}, 61: 178--187.

\bibitem[{Andersson et~al.(2021)Andersson, Nilsson, Shirley, and Akenine{-}M{\"{o}}ller}]{Andersson2021a}
Andersson, P.; Nilsson, J.; Shirley, P.; and Akenine{-}M{\"{o}}ller, T. 2021.
\newblock {Visualizing Errors in Rendered High Dynamic Range Images}.
\newblock In \emph{Eurographics Short Papers}.

\bibitem[{Baltru{\v{s}}aitis, Ahuja, and Morency(2018)}]{baltruvsaitis2018multimodal}
Baltru{\v{s}}aitis, T.; Ahuja, C.; and Morency, L.-P. 2018.
\newblock Multimodal machine learning: A survey and taxonomy.
\newblock \emph{IEEE transactions on pattern analysis and machine intelligence}, 41(2): 423--443.

\bibitem[{Boulahia et~al.(2021)Boulahia, Amamra, Madi, and Daikh}]{boulahia2021early}
Boulahia, S.~Y.; Amamra, A.; Madi, M.~R.; and Daikh, S. 2021.
\newblock Early, intermediate and late fusion strategies for robust deep learning-based multimodal action recognition.
\newblock \emph{Machine Vision and Applications}, 32(6): 121.

\bibitem[{Charbonnier et~al.(1994)Charbonnier, Blanc-Feraud, Aubert, and Barlaud}]{charbonnier1994two}
Charbonnier, P.; Blanc-Feraud, L.; Aubert, G.; and Barlaud, M. 1994.
\newblock Two deterministic half-quadratic regularization algorithms for computed imaging.
\newblock In \emph{Proceedings of 1st international conference on image processing}, volume~2, 168--172. IEEE.

\bibitem[{Chen et~al.(2023{\natexlab{a}})Chen, Liang, Huang, Real, Wang, Pham, Dong, Luong, Hsieh, Lu, and Le}]{NEURIPS2023_9a39b492}
Chen, X.; Liang, C.; Huang, D.; Real, E.; Wang, K.; Pham, H.; Dong, X.; Luong, T.; Hsieh, C.-J.; Lu, Y.; and Le, Q.~V. 2023{\natexlab{a}}.
\newblock Symbolic Discovery of Optimization Algorithms.
\newblock In Oh, A.; Naumann, T.; Globerson, A.; Saenko, K.; Hardt, M.; and Levine, S., eds., \emph{Advances in Neural Information Processing Systems}, volume~36, 49205--49233. Curran Associates, Inc.

\bibitem[{Chen et~al.(2023{\natexlab{b}})Chen, Wang, Zhang, Kong, Qiao, Zhou, and Dong}]{chen2023hat}
Chen, X.; Wang, X.; Zhang, W.; Kong, X.; Qiao, Y.; Zhou, J.; and Dong, C. 2023{\natexlab{b}}.
\newblock Hat: Hybrid attention transformer for image restoration.
\newblock \emph{arXiv preprint arXiv:2309.05239}.

\bibitem[{Chen et~al.(2025)Chen, Nie, Ummenhofer, Birkl, Paulitsch, and Nie{\ss}ner}]{chen2025pbr}
Chen, Y.; Nie, Y.; Ummenhofer, B.; Birkl, R.; Paulitsch, M.; and Nie{\ss}ner, M. 2025.
\newblock PBR-SR: Mesh PBR Texture Super Resolution from 2D Image Priors.
\newblock \emph{arXiv preprint arXiv:2506.02846}.

\bibitem[{Chu et~al.(2024)Chu, Dou, Pan, Weng, and Li}]{Chu_2024_CVPR}
Chu, S.-C.; Dou, Z.-C.; Pan, J.-S.; Weng, S.; and Li, J. 2024.
\newblock HMANet: Hybrid Multi-Axis Aggregation Network for Image Super-Resolution.
\newblock In \emph{Proceedings of the IEEE/CVF Conference on Computer Vision and Pattern Recognition (CVPR) Workshops}, 6257--6266.

\bibitem[{Cook and Torrance(1982)}]{cook1982reflectance}
Cook, R.~L.; and Torrance, K.~E. 1982.
\newblock A reflectance model for computer graphics.
\newblock \emph{ACM Transactions on Graphics (ToG)}, 1(1): 7--24.

\bibitem[{Esser et~al.(2024)Esser, Kulal, Blattmann, Entezari, M{\"u}ller, Saini, Levi, Lorenz, Sauer, Boesel et~al.}]{esser2024scaling}
Esser, P.; Kulal, S.; Blattmann, A.; Entezari, R.; M{\"u}ller, J.; Saini, H.; Levi, Y.; Lorenz, D.; Sauer, A.; Boesel, F.; et~al. 2024.
\newblock Scaling rectified flow transformers for high-resolution image synthesis.
\newblock In \emph{Forty-first international conference on machine learning}.

\bibitem[{Gauthier et~al.(2024)Gauthier, Kerbl, Levallois, Faury, Thiery, and Boubekeur}]{gauthier2024matup}
Gauthier, A.; Kerbl, B.; Levallois, J.; Faury, R.; Thiery, J.-M.; and Boubekeur, T. 2024.
\newblock MatUp: Repurposing image upsamplers for SVBRDFs.
\newblock In \emph{Computer Graphics Forum}, volume~43, e15151. Wiley Online Library.

\bibitem[{Guarrasi et~al.(2025)Guarrasi, Aksu, Caruso, Di~Feola, Rofena, Ruffini, and Soda}]{guarrasi2025systematic}
Guarrasi, V.; Aksu, F.; Caruso, C.~M.; Di~Feola, F.; Rofena, A.; Ruffini, F.; and Soda, P. 2025.
\newblock A systematic review of intermediate fusion in multimodal deep learning for biomedical applications.
\newblock \emph{Image and Vision Computing}, 105509.

\bibitem[{Gunes and Piccardi(2005)}]{fusion_strategy_02}
Gunes, H.; and Piccardi, M. 2005.
\newblock Affect recognition from face and body: early fusion vs. late fusion.
\newblock In \emph{2005 IEEE international conference on systems, man and cybernetics}, volume~4, 3437--3443. IEEE.

\bibitem[{Guo et~al.(2024)Guo, Su, Jiang, Xie, and Liu}]{guo2024cmdaf}
Guo, W.; Su, K.; Jiang, B.; Xie, K.; and Liu, J. 2024.
\newblock CMDAF: Cross-Modality Dual-Attention Fusion Network for Multimodal Sentiment Analysis.
\newblock \emph{Applied Sciences}, 14(24): 12025.

\bibitem[{Ho, Jain, and Abbeel(2020)}]{ho2020denoising}
Ho, J.; Jain, A.; and Abbeel, P. 2020.
\newblock Denoising diffusion probabilistic models.
\newblock \emph{Advances in neural information processing systems}, 33: 6840--6851.

\bibitem[{Hsu, Lee, and Chou(2024)}]{Hsu_2024_CVPR}
Hsu, C.-C.; Lee, C.-M.; and Chou, Y.-S. 2024.
\newblock DRCT: Saving Image Super-Resolution Away from Information Bottleneck.
\newblock In \emph{Proceedings of the IEEE/CVF Conference on Computer Vision and Pattern Recognition (CVPR) Workshops}, 6133--6142.

\bibitem[{Huang et~al.(2017)Huang, Liu, van~der Maaten, and Weinberger}]{DenseNet2017}
Huang, G.; Liu, Z.; van~der Maaten, L.; and Weinberger, K.~Q. 2017.
\newblock Densely connected convolutional networks.
\newblock In \emph{Proceedings of the IEEE Conference on Computer Vision and Pattern Recognition}.

\bibitem[{Johnson, Alahi, and Fei-Fei(2016)}]{johnson2016perceptual}
Johnson, J.; Alahi, A.; and Fei-Fei, L. 2016.
\newblock Perceptual losses for real-time style transfer and super-resolution.
\newblock In \emph{European conference on computer vision}, 694--711. Springer.

\bibitem[{Karis and Games(2013)}]{karis2013real}
Karis, B.; and Games, E. 2013.
\newblock Real shading in unreal engine 4.
\newblock \emph{Proc. Physically Based Shading Theory Practice}, 4(3): 1.

\bibitem[{Lai et~al.(2018)Lai, Huang, Ahuja, and Yang}]{lai2018fast}
Lai, W.-S.; Huang, J.-B.; Ahuja, N.; and Yang, M.-H. 2018.
\newblock Fast and accurate image super-resolution with deep laplacian pyramid networks.
\newblock \emph{IEEE transactions on pattern analysis and machine intelligence}, 41(11): 2599--2613.

\bibitem[{Ledig et~al.(2017{\natexlab{a}})Ledig, Theis, Husz{\'a}r, Caballero, Cunningham, Acosta, Aitken, Tejani, Totz, Wang et~al.}]{ledig2017photo}
Ledig, C.; Theis, L.; Husz{\'a}r, F.; Caballero, J.; Cunningham, A.; Acosta, A.; Aitken, A.; Tejani, A.; Totz, J.; Wang, Z.; et~al. 2017{\natexlab{a}}.
\newblock Photo-realistic single image super-resolution using a generative adversarial network.
\newblock In \emph{Proceedings of the IEEE conference on computer vision and pattern recognition}, 4681--4690.

\bibitem[{Ledig et~al.(2017{\natexlab{b}})Ledig, Theis, Huszár, Caballero, Cunningham, Acosta, Aitken, Tejani, Totz, Wang, and Shi}]{8099502}
Ledig, C.; Theis, L.; Huszár, F.; Caballero, J.; Cunningham, A.; Acosta, A.; Aitken, A.; Tejani, A.; Totz, J.; Wang, Z.; and Shi, W. 2017{\natexlab{b}}.
\newblock Photo-Realistic Single Image Super-Resolution Using a Generative Adversarial Network.
\newblock In \emph{2017 IEEE Conference on Computer Vision and Pattern Recognition (CVPR)}, 105--114.

\bibitem[{Li et~al.(2022)Li, Yang, Chang, Chen, Feng, Xu, Li, and Chen}]{li2022srdiff}
Li, H.; Yang, Y.; Chang, M.; Chen, S.; Feng, H.; Xu, Z.; Li, Q.; and Chen, Y. 2022.
\newblock Srdiff: Single image super-resolution with diffusion probabilistic models.
\newblock \emph{Neurocomputing}, 479: 47--59.

\bibitem[{Li and Tang(2024)}]{fusion_strategy_survey}
Li, S.; and Tang, H. 2024.
\newblock Multimodal Alignment and Fusion: A Survey.
\newblock \emph{arXiv preprint arXiv:2411.17040}.

\bibitem[{Li et~al.(2020)Li, Zhou, Qi, Jiang, Lu, and Jia}]{NEURIPS2020_eaae339c}
Li, W.; Zhou, K.; Qi, L.; Jiang, N.; Lu, J.; and Jia, J. 2020.
\newblock LAPAR: Linearly-Assembled Pixel-Adaptive Regression Network for Single Image Super-resolution and Beyond.
\newblock In Larochelle, H.; Ranzato, M.; Hadsell, R.; Balcan, M.; and Lin, H., eds., \emph{Advances in Neural Information Processing Systems}, volume~33, 20343--20355. Curran Associates, Inc.

\bibitem[{Liang et~al.(2021)Liang, Cao, Sun, Zhang, Van~Gool, and Timofte}]{9607618}
Liang, J.; Cao, J.; Sun, G.; Zhang, K.; Van~Gool, L.; and Timofte, R. 2021.
\newblock SwinIR: Image Restoration Using Swin Transformer.
\newblock In \emph{2021 IEEE/CVF International Conference on Computer Vision Workshops (ICCVW)}, 1833--1844.

\bibitem[{Lim et~al.(2017)Lim, Son, Kim, Nah, and Lee}]{Lim_2017_CVPR_Workshops}
Lim, B.; Son, S.; Kim, H.; Nah, S.; and Lee, K.~M. 2017.
\newblock Enhanced Deep Residual Networks for Single Image Super-Resolution.
\newblock In \emph{The IEEE Conference on Computer Vision and Pattern Recognition (CVPR) Workshops}.

\bibitem[{Liu et~al.(2021)Liu, Lin, Cao, Hu, Wei, Zhang, Lin, and Guo}]{liu2021swin}
Liu, Z.; Lin, Y.; Cao, Y.; Hu, H.; Wei, Y.; Zhang, Z.; Lin, S.; and Guo, B. 2021.
\newblock Swin transformer: Hierarchical vision transformer using shifted windows.
\newblock In \emph{Proceedings of the IEEE/CVF international conference on computer vision}, 10012--10022.

\bibitem[{Pihlgren et~al.(2024)Pihlgren, Nikolaidou, Chhipa, Abid, Saini, Sandin, and Liwicki}]{pihlgren2024systematicperformanceanalysisdeep}
Pihlgren, G.~G.; Nikolaidou, K.; Chhipa, P.~C.; Abid, N.; Saini, R.; Sandin, F.; and Liwicki, M. 2024.
\newblock A Systematic Performance Analysis of Deep Perceptual Loss Networks: Breaking Transfer Learning Conventions.
\newblock arXiv:2302.04032.

\bibitem[{Podell et~al.(2023)Podell, English, Lacey, Blattmann, Dockhorn, M{\"u}ller, Penna, and Rombach}]{podell2023sdxl}
Podell, D.; English, Z.; Lacey, K.; Blattmann, A.; Dockhorn, T.; M{\"u}ller, J.; Penna, J.; and Rombach, R. 2023.
\newblock Sdxl: Improving latent diffusion models for high-resolution image synthesis.
\newblock \emph{arXiv preprint arXiv:2307.01952}.

\bibitem[{Rad et~al.(2019)Rad, Bozorgtabar, Marti, Basler, Ekenel, and Thiran}]{rad2019srobb}
Rad, M.~S.; Bozorgtabar, B.; Marti, U.-V.; Basler, M.; Ekenel, H.~K.; and Thiran, J.-P. 2019.
\newblock Srobb: Targeted perceptual loss for single image super-resolution.
\newblock In \emph{Proceedings of the IEEE/CVF international conference on computer vision}, 2710--2719.

\bibitem[{Saharia et~al.(2022)Saharia, Ho, Chan, Salimans, Fleet, and Norouzi}]{saharia2022image}
Saharia, C.; Ho, J.; Chan, W.; Salimans, T.; Fleet, D.~J.; and Norouzi, M. 2022.
\newblock Image super-resolution via iterative refinement.
\newblock \emph{IEEE transactions on pattern analysis and machine intelligence}, 45(4): 4713--4726.

\bibitem[{Saravanan et~al.(2025)}]{saravanan2025generative}
Saravanan, D.; et~al. 2025.
\newblock A Generative Approach to High Fidelity 3D Reconstruction from Text Data.
\newblock \emph{arXiv preprint arXiv:2503.03664}.

\bibitem[{Schlick(1994)}]{schlick1994inexpensive}
Schlick, C. 1994.
\newblock An inexpensive BRDF model for physically-based rendering.
\newblock In \emph{Computer graphics forum}, volume~13, 233--246. Wiley Online Library.

\bibitem[{Shang et~al.(2024)Shang, Shan, Liu, Wang, Wang, Zhang, and Zhang}]{shang2024resdiff}
Shang, S.; Shan, Z.; Liu, G.; Wang, L.; Wang, X.; Zhang, Z.; and Zhang, J. 2024.
\newblock Resdiff: Combining cnn and diffusion model for image super-resolution.
\newblock In \emph{Proceedings of the AAAI Conference on Artificial Intelligence}, volume~38, 8975--8983.

\bibitem[{Simonyan and Zisserman(2014)}]{simonyan2014very}
Simonyan, K.; and Zisserman, A. 2014.
\newblock Very deep convolutional networks for large-scale image recognition.
\newblock \emph{arXiv preprint arXiv:1409.1556}.

\bibitem[{Snoek, Worring, and Smeulders(2005)}]{fusion_strategy_01}
Snoek, C.~G.; Worring, M.; and Smeulders, A.~W. 2005.
\newblock Early versus late fusion in semantic video analysis.
\newblock In \emph{Proceedings of the 13th annual ACM international conference on Multimedia}, 399--402.

\bibitem[{Sohl-Dickstein et~al.(2015)Sohl-Dickstein, Weiss, Maheswaranathan, and Ganguli}]{sohl2015deep}
Sohl-Dickstein, J.; Weiss, E.; Maheswaranathan, N.; and Ganguli, S. 2015.
\newblock Deep unsupervised learning using nonequilibrium thermodynamics.
\newblock In \emph{International conference on machine learning}, 2256--2265. pmlr.

\bibitem[{Song and Ermon(2019)}]{song2019generative}
Song, Y.; and Ermon, S. 2019.
\newblock Generative modeling by estimating gradients of the data distribution.
\newblock \emph{Advances in neural information processing systems}, 32.

\bibitem[{Song et~al.(2020)Song, Sohl-Dickstein, Kingma, Kumar, Ermon, and Poole}]{song2020score}
Song, Y.; Sohl-Dickstein, J.; Kingma, D.~P.; Kumar, A.; Ermon, S.; and Poole, B. 2020.
\newblock Score-based generative modeling through stochastic differential equations.
\newblock \emph{arXiv preprint arXiv:2011.13456}.

\bibitem[{Trowbridge and Reitz(1975)}]{trowbridge1975average}
Trowbridge, T.; and Reitz, K.~P. 1975.
\newblock Average irregularity representation of a rough surface for ray reflection.
\newblock \emph{Journal of the optical society of America}, 65(5): 531--536.

\bibitem[{Tsai et~al.(2019)Tsai, Bai, Liang, Kolter, Morency, and Salakhutdinov}]{tsai2019multimodal}
Tsai, Y.-H.~H.; Bai, S.; Liang, P.~P.; Kolter, J.~Z.; Morency, L.-P.; and Salakhutdinov, R. 2019.
\newblock Multimodal transformer for unaligned multimodal language sequences.
\newblock In \emph{Proceedings of the conference. Association for computational linguistics. Meeting}, volume 2019, 6558.

\bibitem[{Vaswani et~al.(2017)Vaswani, Shazeer, Parmar, Uszkoreit, Jones, Gomez, Kaiser, and Polosukhin}]{vaswani2017attention}
Vaswani, A.; Shazeer, N.; Parmar, N.; Uszkoreit, J.; Jones, L.; Gomez, A.~N.; Kaiser, {\L}.; and Polosukhin, I. 2017.
\newblock Attention is all you need.
\newblock \emph{Advances in neural information processing systems}, 30.

\bibitem[{Vecchio and Deschaintre(2024)}]{vecchio2024matsynth}
Vecchio, G.; and Deschaintre, V. 2024.
\newblock Matsynth: A modern pbr materials dataset.
\newblock In \emph{Proceedings of the IEEE/CVF Conference on Computer Vision and Pattern Recognition}, 22109--22118.

\bibitem[{Vecchio et~al.(2024{\natexlab{a}})Vecchio, Martin, Roullier, Kaiser, Rouffet, Deschaintre, and Boubekeur}]{vecchio2024controlmat}
Vecchio, G.; Martin, R.; Roullier, A.; Kaiser, A.; Rouffet, R.; Deschaintre, V.; and Boubekeur, T. 2024{\natexlab{a}}.
\newblock Controlmat: a controlled generative approach to material capture.
\newblock \emph{ACM Transactions on Graphics}, 43(5): 1--17.

\bibitem[{Vecchio et~al.(2024{\natexlab{b}})Vecchio, Sortino, Palazzo, and Spampinato}]{vecchio2024matfuse}
Vecchio, G.; Sortino, R.; Palazzo, S.; and Spampinato, C. 2024{\natexlab{b}}.
\newblock Matfuse: controllable material generation with diffusion models.
\newblock In \emph{Proceedings of the IEEE/CVF Conference on Computer Vision and Pattern Recognition}, 4429--4438.

\bibitem[{Wang et~al.(2018)Wang, Yu, Wu, Gu, Liu, Dong, Qiao, and Loy}]{wang2018esrgan}
Wang, X.; Yu, K.; Wu, S.; Gu, J.; Liu, Y.; Dong, C.; Qiao, Y.; and Loy, C.~C. 2018.
\newblock ESRGAN: Enhanced super-resolution generative adversarial networks.
\newblock In \emph{The European Conference on Computer Vision Workshops (ECCVW)}.

\bibitem[{Wang et~al.(2024)Wang, Yang, Chen, Wang, Guo, Chau, Liu, Qiao, Kot, and Wen}]{wang2024sinsr}
Wang, Y.; Yang, W.; Chen, X.; Wang, Y.; Guo, L.; Chau, L.-P.; Liu, Z.; Qiao, Y.; Kot, A.~C.; and Wen, B. 2024.
\newblock Sinsr: diffusion-based image super-resolution in a single step.
\newblock In \emph{Proceedings of the IEEE/CVF conference on computer vision and pattern recognition}, 25796--25805.

\bibitem[{Wu et~al.(2017)Wu, Duan, Liu, and Sun}]{wu2017srpgan}
Wu, B.; Duan, H.; Liu, Z.; and Sun, G. 2017.
\newblock SRPGAN: perceptual generative adversarial network for single image super resolution.
\newblock \emph{arXiv preprint arXiv:1712.05927}.

\bibitem[{Xiao et~al.(2021)Xiao, Singh, Mintun, Darrell, Dollar, and Girshick}]{NEURIPS2021_ff1418e8}
Xiao, T.; Singh, M.; Mintun, E.; Darrell, T.; Dollar, P.; and Girshick, R. 2021.
\newblock Early Convolutions Help Transformers See Better.
\newblock In Ranzato, M.; Beygelzimer, A.; Dauphin, Y.; Liang, P.; and Vaughan, J.~W., eds., \emph{Advances in Neural Information Processing Systems}, volume~34, 30392--30400. Curran Associates, Inc.

\bibitem[{Xu et~al.(2015)Xu, Wang, Chen, and Li}]{xu2015empirical}
Xu, B.; Wang, N.; Chen, T.; and Li, M. 2015.
\newblock Empirical evaluation of rectified activations in convolutional network.
\newblock \emph{arXiv preprint arXiv:1505.00853}.

\bibitem[{Ye et~al.(2024)Ye, Qiu, Gu, Zuo, Wu, Dong, Bo, Xiu, and Han}]{ye2024stablenormal}
Ye, C.; Qiu, L.; Gu, X.; Zuo, Q.; Wu, Y.; Dong, Z.; Bo, L.; Xiu, Y.; and Han, X. 2024.
\newblock Stablenormal: Reducing diffusion variance for stable and sharp normal.
\newblock \emph{ACM Transactions on Graphics (TOG)}, 43(6): 1--18.

\bibitem[{Yue, Liao, and Loy(2025)}]{yue2025arbitrary}
Yue, Z.; Liao, K.; and Loy, C.~C. 2025.
\newblock Arbitrary-steps image super-resolution via diffusion inversion.
\newblock In \emph{Proceedings of the Computer Vision and Pattern Recognition Conference}, 23153--23163.

\bibitem[{Yue, Wang, and Loy(2023)}]{yue2023resshift}
Yue, Z.; Wang, J.; and Loy, C.~C. 2023.
\newblock Resshift: Efficient diffusion model for image super-resolution by residual shifting.
\newblock \emph{Advances in Neural Information Processing Systems}, 36: 13294--13307.

\bibitem[{Zhang et~al.(2020)Zhang, Fan, Dai, Anwar, Saleh, Zhang, and Barnes}]{zhang2020uc}
Zhang, J.; Fan, D.-P.; Dai, Y.; Anwar, S.; Saleh, F.~S.; Zhang, T.; and Barnes, N. 2020.
\newblock UC-Net: Uncertainty inspired RGB-D saliency detection via conditional variational autoencoders.
\newblock In \emph{Proceedings of the IEEE/CVF conference on computer vision and pattern recognition}, 8582--8591.

\bibitem[{Zhang et~al.(2023)Zhang, Lei, Xie, Fang, Li, and Du}]{zhang2023superyolo}
Zhang, J.; Lei, J.; Xie, W.; Fang, Z.; Li, Y.; and Du, Q. 2023.
\newblock SuperYOLO: Super resolution assisted object detection in multimodal remote sensing imagery.
\newblock \emph{IEEE Transactions on Geoscience and Remote Sensing}, 61: 1--15.

\bibitem[{Zhang et~al.(2021)Zhang, Liang, Van~Gool, and Timofte}]{zhang2021designing}
Zhang, K.; Liang, J.; Van~Gool, L.; and Timofte, R. 2021.
\newblock Designing a practical degradation model for deep blind image super-resolution.
\newblock In \emph{Proceedings of the IEEE/CVF international conference on computer vision}, 4791--4800.

\bibitem[{Zhang et~al.(2018{\natexlab{a}})Zhang, Isola, Efros, Shechtman, and Wang}]{zhang2018unreasonable}
Zhang, R.; Isola, P.; Efros, A.~A.; Shechtman, E.; and Wang, O. 2018{\natexlab{a}}.
\newblock The unreasonable effectiveness of deep features as a perceptual metric.
\newblock In \emph{Proceedings of the IEEE conference on computer vision and pattern recognition}, 586--595.

\bibitem[{Zhang et~al.(2018{\natexlab{b}})Zhang, Li, Li, Wang, Zhong, and Fu}]{zhang2018image}
Zhang, Y.; Li, K.; Li, K.; Wang, L.; Zhong, B.; and Fu, Y. 2018{\natexlab{b}}.
\newblock Image super-resolution using very deep residual channel attention networks.
\newblock In \emph{Proceedings of the European conference on computer vision (ECCV)}, 286--301.

\bibitem[{Zhao et~al.(2020)Zhao, Zhang, Pang, Lu, and Zhang}]{zhao2020single}
Zhao, X.; Zhang, L.; Pang, Y.; Lu, H.; and Zhang, L. 2020.
\newblock A single stream network for robust and real-time RGB-D salient object detection.
\newblock In \emph{Computer Vision--ECCV 2020: 16th European Conference, Glasgow, UK, August 23--28, 2020, Proceedings, Part XXII 16}, 646--662. Springer.

\bibitem[{Zhou et~al.(2023)Zhou, Li, Guo, Bai, Cheng, and Hou}]{zhou2023srformer}
Zhou, Y.; Li, Z.; Guo, C.-L.; Bai, S.; Cheng, M.-M.; and Hou, Q. 2023.
\newblock Srformer: Permuted self-attention for single image super-resolution.
\newblock In \emph{Proceedings of the IEEE/CVF international conference on computer vision}, 12780--12791.

\end{thebibliography}

\setlength{\leftmargini}{20pt}
\makeatletter\def\@listi{\leftmargin\leftmargini \topsep .5em \parsep .5em \itemsep .5em}
\def\@listii{\leftmargin\leftmarginii \labelwidth\leftmarginii \advance\labelwidth-\labelsep \topsep .4em \parsep .4em \itemsep .4em}
\def\@listiii{\leftmargin\leftmarginiii \labelwidth\leftmarginiii \advance\labelwidth-\labelsep \topsep .4em \parsep .4em \itemsep .4em}\makeatother

\setcounter{secnumdepth}{0}
\renewcommand\thesubsection{\arabic{subsection}}
\renewcommand\labelenumi{\thesubsection.\arabic{enumi}}

\newcounter{checksubsection}
\newcounter{checkitem}[checksubsection]

\newcommand{\checksubsection}[1]{%
  \refstepcounter{checksubsection}%
  \paragraph{\arabic{checksubsection}. #1}%
  \setcounter{checkitem}{0}%
}

\newcommand{\checkitem}{%
  \refstepcounter{checkitem}%
  \item[\arabic{checksubsection}.\arabic{checkitem}.]%
}
\newcommand{\question}[2]{\normalcolor\checkitem #1 #2 \color{blue}}
\newcommand{\ifyespoints}[1]{\makebox[0pt][l]{\hspace{-15pt}\normalcolor #1}}

\newpage

\end{document}